\documentclass[11pt]{article}

\usepackage[final]{acl}
\newcommand\blfootnote[1]{%
  \begingroup
  \renewcommand\thefootnote{}\footnote{#1}%
  \addtocounter{footnote}{-1}%
  \endgroup
}
\usepackage{times}
\usepackage{latexsym}
\usepackage{algorithm}
\usepackage{algpseudocode}
\usepackage{algorithmicx}
\usepackage{amsmath}
\usepackage{amssymb}
\usepackage{multirow}
\usepackage[table]{xcolor}
\usepackage{tabularx}
\usepackage{arydshln}   % 用于画虚线 (\hdashline)
\usepackage{subcaption}  % 导入 subcaption 包
\usepackage{pifont}

\newcolumntype{Y}{>{\centering\arraybackslash}X}
\usepackage{booktabs}
\usepackage[T1]{fontenc}
\usepackage[utf8]{inputenc}

\usepackage{microtype}

\usepackage{inconsolata}

\usepackage{graphicx}

\title{Selective Contrastive Learning For Gloss Free Sign Language Translation}

\author{
    Changhao Lai\textsuperscript{1,2,3$^{*}$},
    Rui Zhao\textsuperscript{1,2,3$^{*}$},
    Xuewen Zhong\textsuperscript{1,2,3}, \\
    \textbf{Jinsong Su\textsuperscript{1,2}} \and
    \textbf{Yidong Chen\textsuperscript{1,2,3$^{\dagger}$}} \\
    $^1$School of Informatics, Xiamen University, China \\
    $^2$Key Lab of Digital Protection and Intelligent Processing of Intangible \\ Cultural Heritage of Fujian-Taiwan (XMU), Ministry of Culture and Tourism, China \\
    $^3$National Language Resources Monitoring and \\ Research Center for Education and Teaching Media, Xiamen University, China \\
    \texttt{laichanghao@stu.xmu.edu.cn \quad ydchen@xmu.edu.cn} \\
}

\begin{document}
\maketitle

% \blfootnote{\textsuperscript{*}Equal contribution.}
% \blfootnote{\textsuperscript{$\dagger$}\textbf{Corresponding Author.} 
% \href{mailto:ydchen@xmu.edu.cn}{ydchen@xmu.edu.cn}}

\blfootnote{\textsuperscript{*}Equal contribution. }
\blfootnote{\textsuperscript{$\dagger$}Corresponding Author.} 
% \href{mailto:ydchen@xmu.edu.cn}{ydchen@xmu.edu.cn}}

\begin{abstract}

Sign language translation (SLT) converts continuous sign videos into spoken-language text, yet it remains challenging due to the intrinsic modality mismatch between visual signs and written text, particularly in gloss-free settings. 
Recent SLT systems increasingly adopt CLIP-like Vision-Language pretraining (VLP) for cross-modal alignment, but the random in-batch contrast provides few, batch-dependent negatives and may mislabel semantically similar (or even identical) pairs as negatives, introducing noisy and potentially inconsistent alignment supervision.
In this work, we first conduct a preliminary trajectory-based analysis that tracks negative video-text similarity over training. The results show that only a small subset of negatives exhibits the desired behavior of being consistently pushed away, while the remaining negatives display heterogeneous and often non-decreasing similarity dynamics, suggesting that random in-batch negatives are frequently uninformative for effective alignment.
Inspired by this, we propose Selective Contrastive Learning for SLT (SCL-SLT) with a Pair Selection (PS) strategy. PS scores candidate negatives using similarity dynamics from reference checkpoints and constructs mini-batches via a curriculum that progressively emphasizes more challenging negatives, thereby strengthening contrastive supervision while reducing the influence of noisy or semantically invalid negatives. 
% Extensive experiments on PHOENIX14T and CSL-Daily demonstrate that SCL-SLT significantly outperforms vanilla contrastive learning across different training settings.
% SCL-SLT improves over vanilla contrastive learning with random in-batch negatives, under end-to-end training, SCL-SLT achieves BLEU scores of 25.30 (+3.27) on PHOENIX14T and 21.41 (+0.82) on CSL-Daily; with standard VLP, it achieves 26.00 (+0.99) on PHOENIX14T and 23.25 (+2.55) on CSL-Daily.
% On PHOENIX14T and CSL-Daily, SCL-SLT improves over vanilla contrastive learning with random in-batch negatives, achieving  BLEU scores of 24.59/21.41 under end-to-end training and 25.98/23.25 with standard VLP. 

\end{abstract}

\begin{figure}[t]
  \includegraphics[width=\columnwidth]{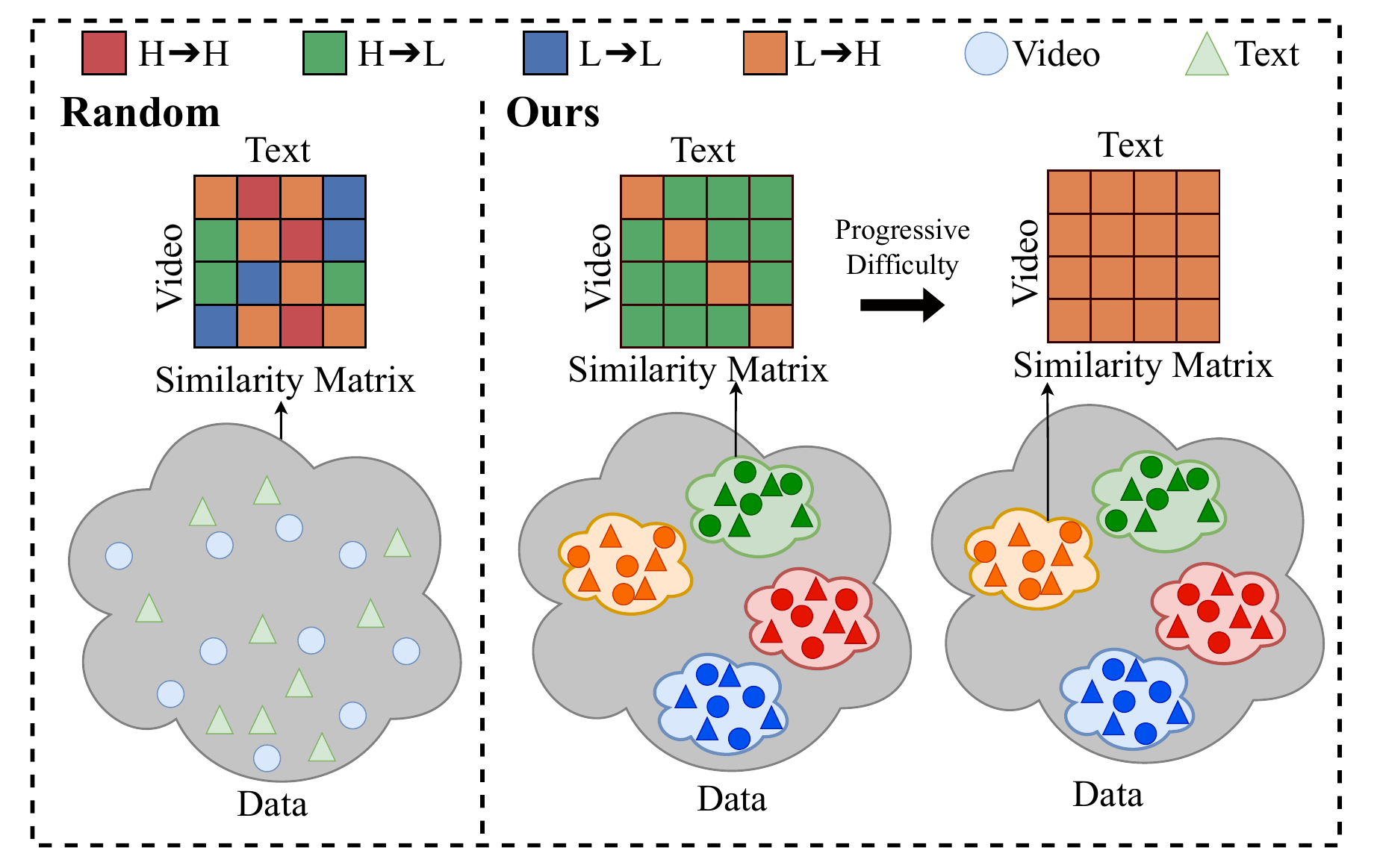}
  %\caption{批次构建策略与数据特征的概念性对比。\textbf{(上图)} 随机批次构建（左）与我们提出的动态样本对选择机制（右）的对比。不同于随机采样会导致相似度矩阵杂乱无序且学习信号混杂，我们的方法通过策略性地选择样本对来引入“渐进式难度 (Progressive Difficulty)”，将批次按由易到难的方式组织，从而优化对比学习的动态过程。\textbf{(下图)} 来自 PHOENIX14T（左）和 CSL-Daily（右）数据集的示例。这些样本展示了手语数据中固有的高结构相似性和词汇重叠现象，这也阐明了采用我们提出的选择性方法而非随机采样的必要性。}
\caption{
% Conceptual comparison of batch construction strategies. Unlike random sampling (Left) which yields disordered similarity matrices and mixed difficulty levels, our dynamic selection (Right) introduces ``Progressive Difficulty'', organizing batches from easy to hard. 
Comparison between the vanilla contrastive learning (left) and our Selective Contrastive Learning (right). SCL progressively selects highly informative negatives for efficient and effective alignment in SLT.
}
% The legends denote the similarity trend of a pair during training: HH (remains high), HL (decreases), LL (remains low), and LH (increases).
\label{fig:method-simple}
\end{figure}

\section{Introduction}

Sign language serves as a primary means of communication for the Deaf and hard-of-hearing community worldwide.
As a vision-centric language, it possesses the full range of fundamental linguistic properties, with meaning expressed through the coordinated use of manual and non-manual articulatory cues, e.g., hand shapes, movements, and facial expressions. These characteristics present significant challenges for automatic sign language understanding and have simultaneously drawn growing attention from the research community, particularly in sign language translation (SLT), which aims to translate continuous sign video sequences into spoken language text~\cite{PHOENIX14T, gloss-SLT-SLRT, gloss-SLT-BN-TIN-Transf, gloss-SLT-MMTLB, gloss-free-VLP, fuSignerDiversitydrivenData2024a, zhaoConditionalVariationalAutoencoder2024, gloss-free-MMSLT, zhangDynamicFeatureFusion2025, fuImprovingEndtoEndSign2025}.

\begin{figure}[t]
\centering
\includegraphics[width=\linewidth]{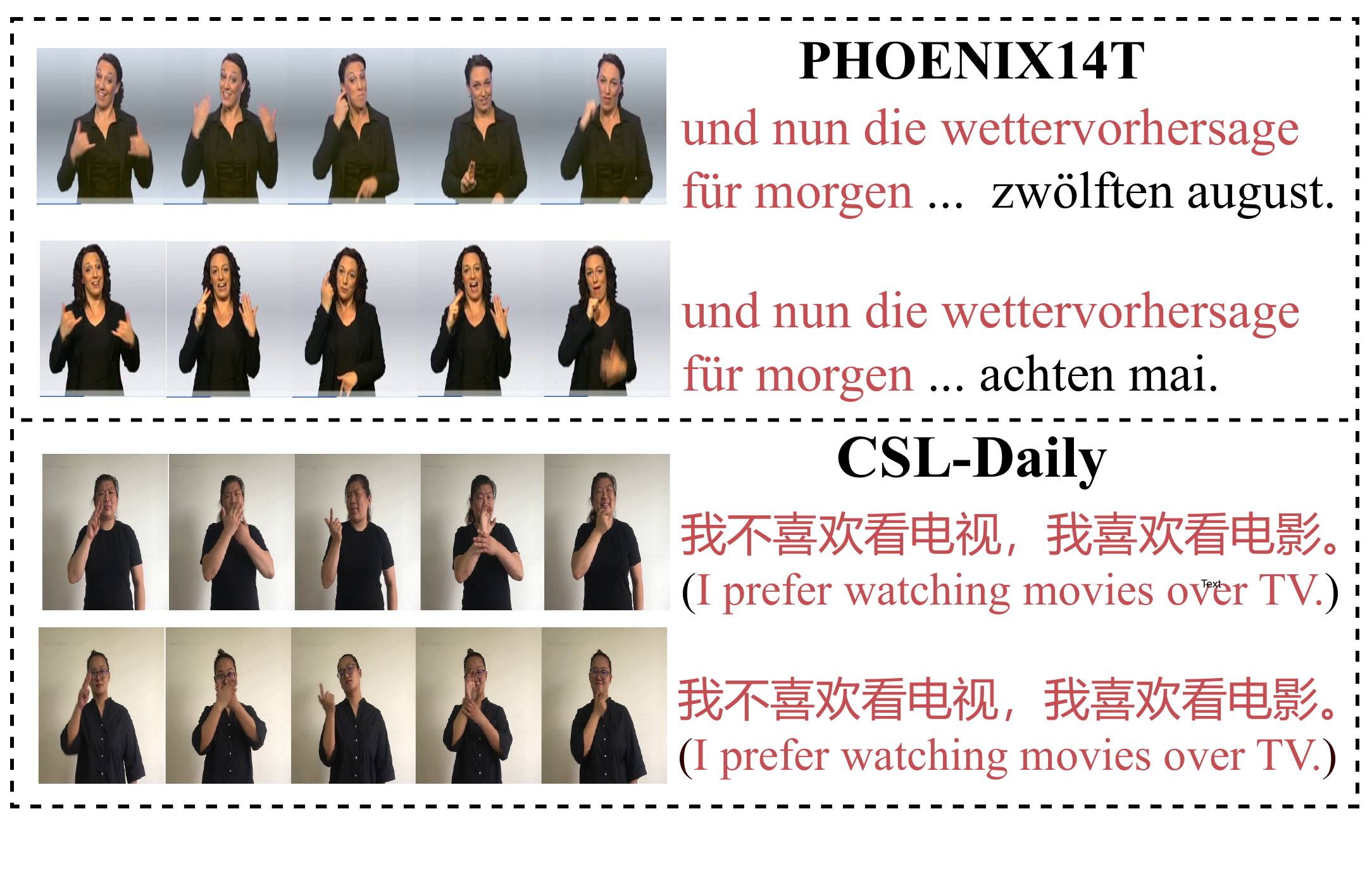}
% \vspace{2}
% \caption{Examples of semantic redundancy.}
\caption{Semantically similar or identical instances.}
\label{fig_semantic_similar_cases}
% \caption{(a) Examples from PHOENIX14T and CSL-Daily exhibit high structural similarity and lexical overlap, highlighting the inadequacy of random sampling. (b) The similarity trajectories of four negative pair categories. The statistics reveal that only $35.9\%$ of negative samples (H~$\rightarrow$~L) behave as expected.
% % while other negatives contradict the contrastive learning objective.
% }
\label{fig_cases_in_phe_csl}
\end{figure}

However, SLT inherently faces a substantial modality mismatch between sign videos and written text, which remains a key bottleneck for translation quality and keeps current performance suboptimal.
This challenge is further amplified in gloss-free\footnote{Glosses are word-level spoken-language annotations that roughly correspond to sign meanings, typically transcribed in a fixed sign-by-sign order.} SLT (GFSLT), where gloss annotations are unavailable, and the model is forced to learn the video-to-text mapping without an intermediate supervision for alignment.
To mitigate this gap, a prominent line of work adopts CLIP-like Vision-Language pretraining (VLP) to strengthen cross-modal alignment, yielding consistent empirical gains in recent SLT systems~\cite{gloss-free-VLP, gloss-free-cico, gloss-free-signcl, gloss-free-llavaslt, gloss-free-c2rl}. 
As illustrated in Figure~\ref{fig:method-simple}~(left), these methods learn a shared embedding space through mini-batch contrast, pulling each matched video--text pair closer while pushing it away from all other non-matching pairs within the same mini-batch. 
Despite its effectiveness, this in-batch design provides each example with only a limited and batch-dependent set of negative pairs per update, and the problem is further exacerbated by the high computational cost associated with spatio-temporal video modeling, which often limits the achievable batch size. 
More critically, in-batch contrastive learning implicitly assumes that negatives are semantically distinct from the anchor. In practice, however, semantically similar or even semantically identical samples may be treated as negatives, as exemplified in Figure~\ref{fig_cases_in_phe_csl}, potentially introducing conflicting supervision signals and impeding reliable cross-modal alignment.

\begin{figure}[t]
\centering
\includegraphics[width=\linewidth]{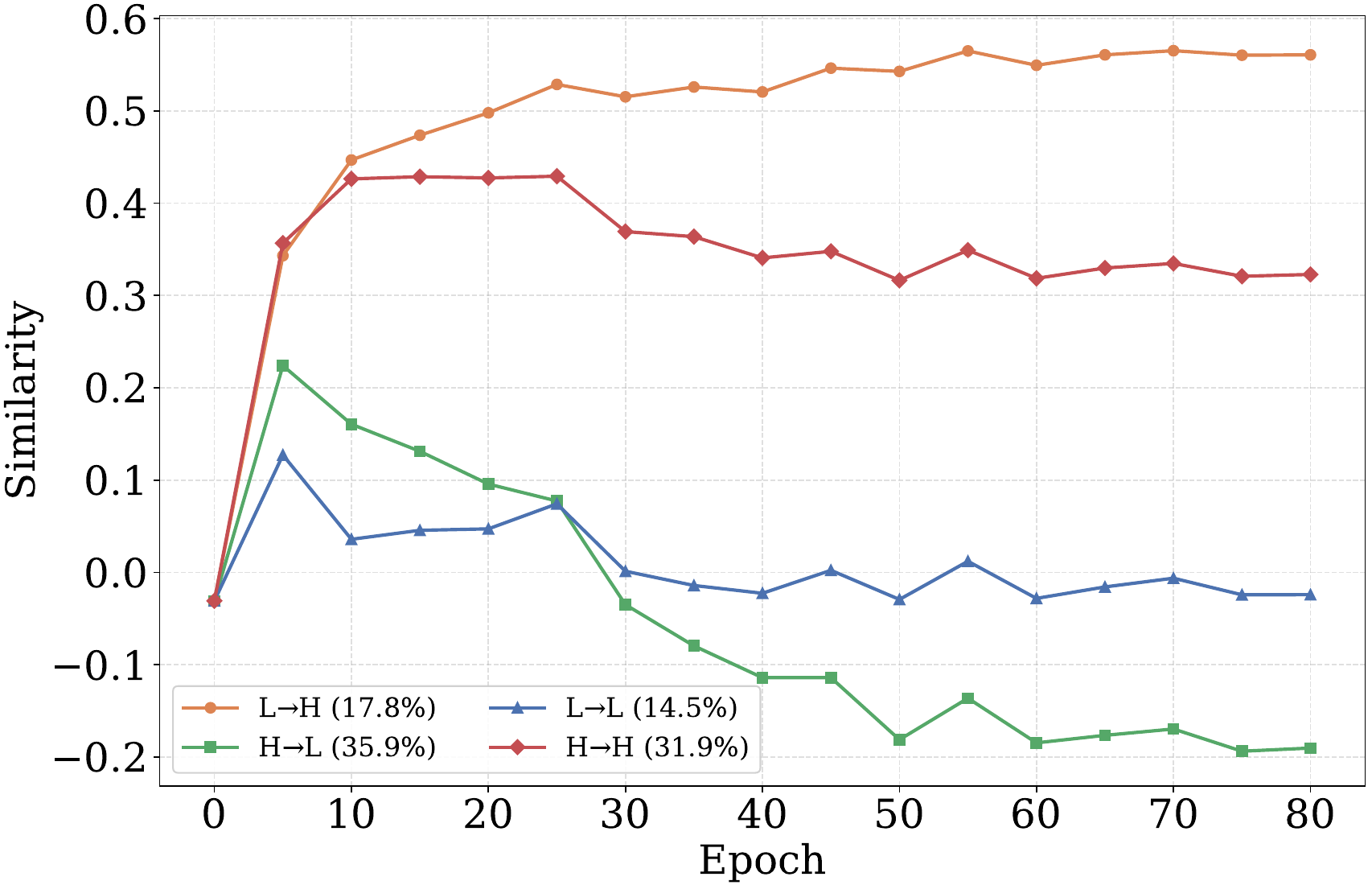}
% \caption{The four categories in in-batch negatives.}
\caption{Similarity trajectories of four categories.}
\label{fig_four_subsets}
% \caption{(a) Examples from PHOENIX14T and CSL-Daily exhibit high structural similarity and lexical overlap, highlighting the inadequacy of random sampling. (b) The similarity trajectories of four negative pair categories. The statistics reveal that only $35.9\%$ of negative samples (H~$\rightarrow$~L) behave as expected.
% % while other negatives contradict the contrastive learning objective.
% }
\label{fig_class_trajectory}
\end{figure}

\begin{figure*}[t]
\centering
\includegraphics[width=\textwidth]{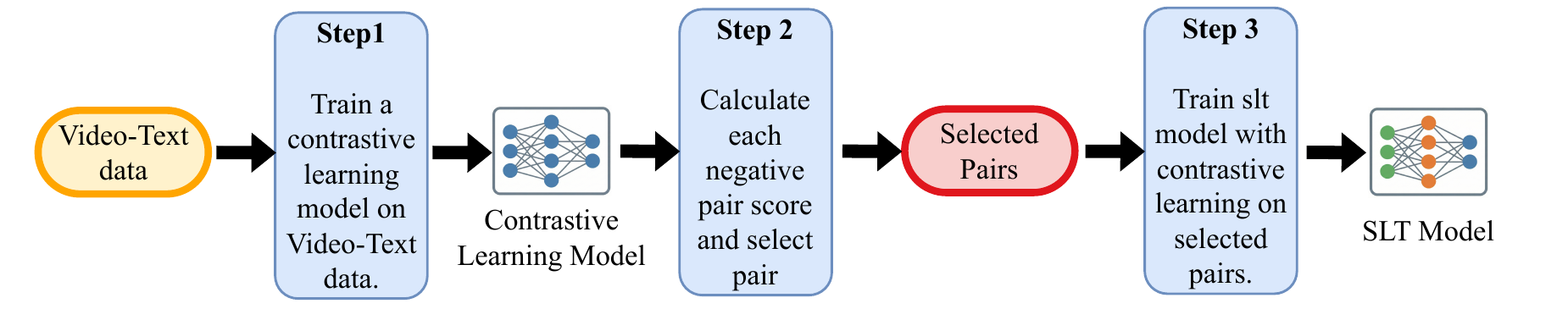}  % 使用 \textwidth
\caption{\textbf{Overview of the SCL-SLT pipeline.} The process consists of three stages: (Step 1) Training a preliminary contrastive learning model on video-text data, (Step 2) Computing similarity scores to select informative negative pairs, and (Step 3) Fine-tuning the target SLT model using contrastive learning on the selected pairs.}
\label{fig:pipeline}
\end{figure*}

To address this limitation, we conduct a preliminary study that tracks how the similarity of negative video-text pairs evolves over training (Section~\ref{sec_preliminary}). 
The results in Figure~\ref{fig_class_trajectory} reveal that only a small fraction of negatives provide informative contrastive supervision, in the sense that their similarity is consistently reduced over training (H~$\rightarrow$~L). Most negatives instead fall into two broad regimes: \emph{easy} pairs that remain consistently dissimilar (L~$\rightarrow$~L), and \emph{hard} pairs that stay highly similar with substantial fluctuations (H~$\rightarrow$~H), while a smaller portion exhibits increasingly challenging similarity over training (L~$\rightarrow$~H).
Building on these observations, we propose \textbf{S}elective \textbf{C}ontrastive \textbf{L}earning for SLT (\textbf{SCL-SLT}) with a Pair Selection (PS) strategy (Section~\ref{subsec:PS}). As shown in Figure~\ref{fig:method-simple} (right), PS dynamically prioritizes informative negative pairs to strengthen the contrastive supervision signal. 
The overall pipeline (Figure~\ref{fig:pipeline}) consists of three stages. We first train a standard contrastive baseline with in-batch random negatives as a reference model. Then, checkpoints of the reference model are collected during reference training to score candidate negatives by their similarity trajectories (e.g., the change in similarity from early to late training), and retain highly informative negatives accordingly. 
% Finally, we adopt a curriculum strategy to progressively shift the selected negatives from \emph{easy} pairs to \emph{hard} pairs during SLT training, which corresponds to increasingly challenging similarity dynamics.
Finally, we adopt a curriculum strategy to progressively shift the selected negatives from \emph{easy} to \emph{hard} during SLT training, moving from negatives that are already low-similarity or readily separable to those that remain highly similar or become increasingly similar. 
% Extensive experimental results demonstrate the superiority of our proposed approach. Compared to the baseline employing a random selection strategy (CL-SLT), our SCL-SLT achieves a significant 2.56 BLEU-4 improvement on PHOENIX14T. Notably, it reaches 24.59 BLEU-4 even without any fine-tuning, a result surpassed only by the current SOTA (MMSLT) and our own fine-tuned model, thus validating the intrinsic effectiveness of the proposed SCL strategy. Furthermore, with subsequent fine-tuning, SCL-SLT-F establishes new state-of-the-art (SOTA) performance on the two most widely used SLT benchmarks, PHOENIX14T and CSL-Daily.
% Empirically, SCL-SLT yields consistent gains on two widely used SLT benchmarks over vanilla contrastive learning. 
% Under End-to-End settings, the model attains 24.59 and 21.41 BLEU scores on PHOENIX14T and CSL-Daily, respectively, an improvement of +2.56 and +0.82 over the random in-batch negatives contrastive learning. 
% With standard Vision-Language Pretraining, the resulting model reaches 25.98 and 23.25 scores, achieving a new state-of-the-art. 

\begin{figure*}[ht]
\centering
\begin{subfigure}{0.48\linewidth}
    \centering
    \includegraphics[width=\linewidth]{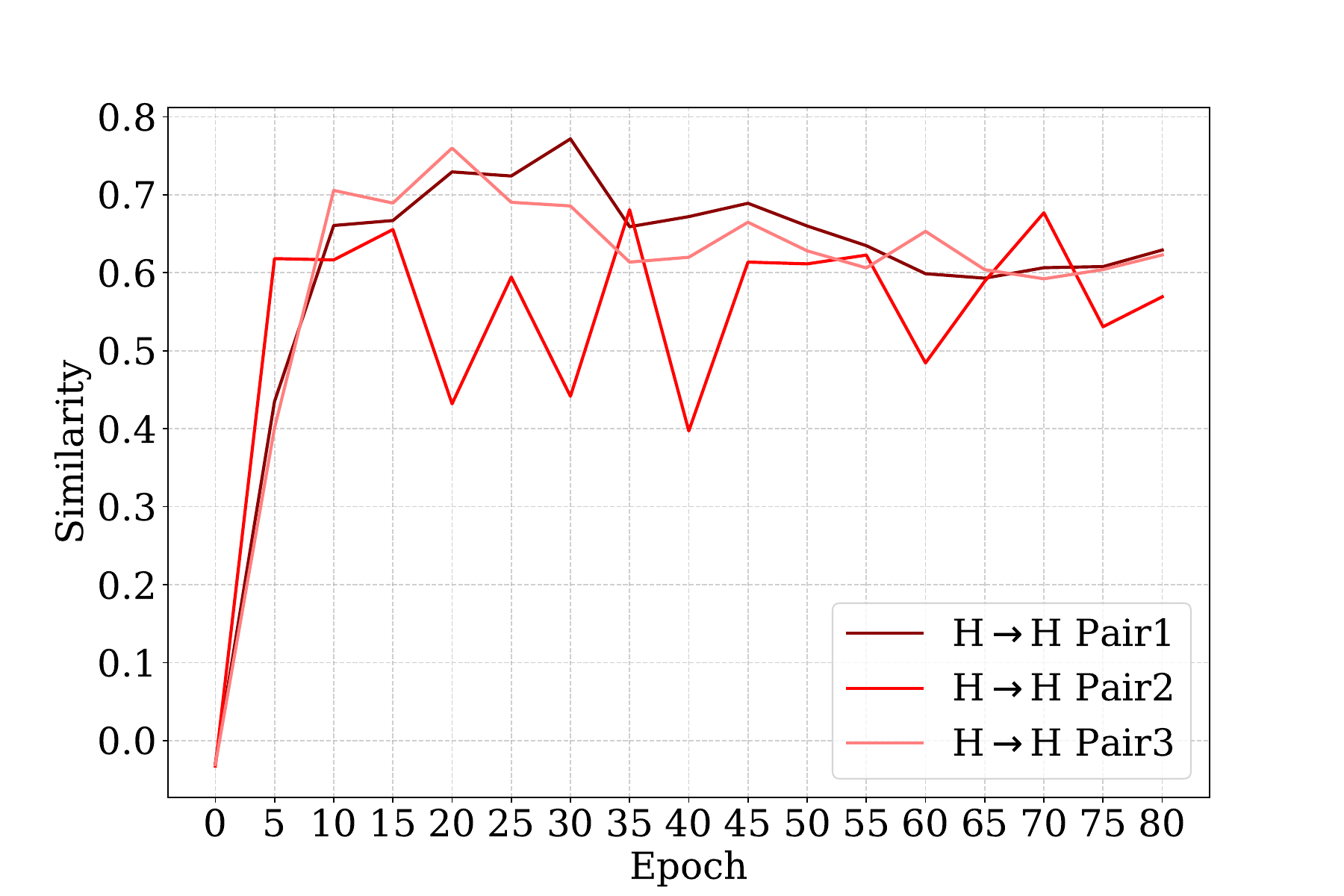}
    % \vspace{-2mm}
    % \caption{H~$\rightarrow$~H: High-Stable Similarity}
    \caption{Example of negatives in H~$\rightarrow$~H.}
    \label{fig:hh_case}
\end{subfigure}
\hfill
\begin{subfigure}{0.48\linewidth}
    \centering
    \includegraphics[width=\linewidth]{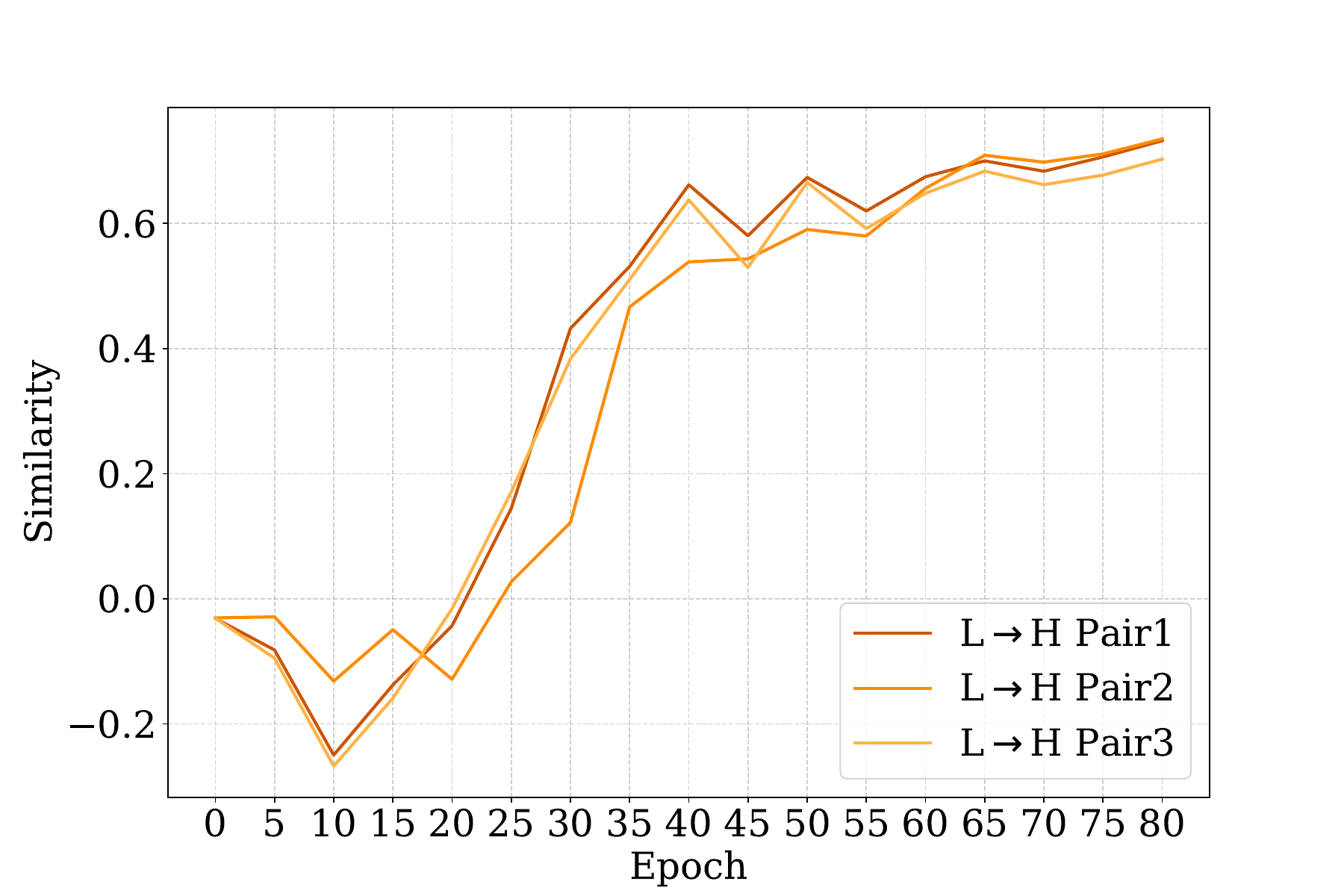}
    \caption{Example of negatives in L~$\rightarrow$~H.}
    \label{fig:lh_case}
\end{subfigure}
\caption{
% Evolution of similarity scores for representative negative pairs during training. 
% Similarity trajectories of representative negatives, 
(a) The negative pairs in H~$\rightarrow$~H exhibit increasing similarity over training,
(b) The negatives in L~$\rightarrow$~H show high similarity with fluctuations. Both cases demonstrate resistance to distinction during contrastive learning.
}\label{fig:similarity_case}
\end{figure*}

% % ========================= add positioning =========================
Unlike recent methods that heavily rely on massive Language Models (LLMs) or complex auxiliary annotations, SCL-SLT takes an orthogonal, data-centric approach. We demonstrate that rigorously optimizing intrinsic contrastive signals alone is fully sufficient to achieve superior translation precision, without introducing extraneous computational overhead or external data dependencies.

In summary, our contributions are as follows:

(1) We present a similarity trajectory analysis of in-batch negative pairs for CLIP-like contrastive learning in SLT, demonstrating that negatives contribute unevenly and impede reliable alignment.

(2) We propose SCL-SLT, which prioritizes informative negatives while mitigating the impact of noisy or semantically invalid negatives via a curriculum-guided Pair Selection mechanism. 

% (3) The experiments on PHOENIX14T and CSL-Daily show that SCL-SLT consistently outperforms vanilla contrastive learning with random in-batch negatives, under end-to-end training, SCL-SLT achieves BLEU scores of 25.30 (+3.27) on PHOENIX14T and 21.41 (+0.82) on CSL-Daily; with standard VLP, it achieves 26.00 (+0.99) on PHOENIX14T and 23.25 (+2.55) on CSL-Daily.
(3) Extensive experiments on PHOENIX14T and CSL-Daily demonstrate that SCL-SLT significantly outperforms vanilla contrastive learning.

% % ========================= introduction end =========================

\section{Preliminary}\label{sec_preliminary}
For an SLT dataset $D={\{V_i, T_i\}}_{i=1}^N$ containing $N$ paired sign video-text instances, the goal is to translate each source video $V_i$ into its target sentence $T_i$.
Due to the inherent nature of sign language and SLT maps sign videos to written text, the modality mismatch between the two remains a key challenge and often limits translation quality. 
Accordingly, a growing body of work adopts CLIP-like VLP to strengthen video-text alignment.

\paragraph{Limitations of In-Batch Negatives.}\label{sec:pre-experiment}
% Given a mini-batch $\mathcal{B}={\{Vi, Ti\}}_{i=1}^{B}$ sampled from the training data, CLIP-like training treats each paired $(V_i, T_i)$ as a positive sample, while $(V_i, T_j)$ for $i \ne j$ form in-batch negatives.
Given a mini-batch $\mathcal{B}={\{V_i, T_i\}}_{i=1}^{B}$ sampled from the training data, CLIP-like training treats each paired $(V_i, T_i)$ as a positive, while pairing $V_i$ with other texts $ T_j(i \ne j)$ yields in-batch negatives. 
% A similarity matrix $S\in \mathbb{R}^{B \times B}$ is computed between all video and text representations within the batch, where diagonal entries correspond to positives and off-diagonal entries correspond to negatives. 
The resulting similarity matrix $S\in \mathbb{R}^{B \times B}$ places positives on the diagonal and negatives on the off-diagonal entries.
The contrastive objective then pulls matched pairs together and pushes mismatched pairs apart in a shared embedding space (Figure~\ref{fig:method-simple}, right). 
% Under this in-batch design, each video is contrasted against only $|B| - 1$ negatives per update, and the negative set that participates in training is determined by the random composition of mini-batches, resulting in a negative coverage ratio of $(B-1)/(N-1)$, and many potentially informative negatives are rarely encountered during training since $B << N$.
Under this in-batch design, each video is contrasted against only $B-1$ negatives per update, and the participating negatives depend on the random mini-batch composition. Consequently, the per-update negative coverage is $(B-1)/(N-1)$, which can be very small when $B \ll N$, thus many potentially \textit{informative negatives are seldom observed in typical training.} 
This issue is further exacerbated by the high computational cost of spatio-temporal video modeling, which often necessitates smaller batch sizes and consequently limits the number and diversity of in-batch negatives. 
More importantly, the common assumption that \textit{in-batch negatives are semantically distinct may not hold in SLT corpora}: the PHOENIX14~\cite{PHOENIX14T} includes many pairs with substantially similar semantics, and CSL-Daily~\cite{gloss-SLT-BN-TIN-Transf} further contains numerous pairs with identical target sentences (Figure~\ref{fig_semantic_similar_cases}). 
Therefore, semantically similar or even identical pairs can be treated as negatives (false negatives), producing inconsistent supervision signals that can hinder robust cross-modal alignment. These observations motivate a more principled treatment of negatives that accounts for both coverage and semantic validity, which is the focus of our approach.

\paragraph{Quantitative Study of In-Batch Negatives.} 
% To study how  evolve during training
To investigate how negative video-text similarities evolve over training, we train a contrastive model on the PHOENIX14T following the GFSLT-VLP framework~\cite{gloss-free-VLP} and track the similarity changes of each paired sample at intervals of 5 epochs.
% Specifically, checkpoints are saved at intervals of 5 epochs, and the representations of sign videos and corresponding target texts in the complete training set are extracted from each checkpoint to compute the similarity matrices according to Equation~\ref{eq:batch_score}. 
A clear pattern emerges from the similarity trajectories. while the diagonal elements (positive) align perfectly with the goal of contrastive learning, the off-diagonal elements (negatives) exhibit substantial heterogeneity. It suggests that while positives are effectively pulled closer, \textit{not all negatives are successfully pushed apart.}
We further quantitatively analyze the changes in similarity of these negatives using a linear least-squares regression. 
Finally, we stratify these negatives into four distinct categories based on their fitted trends: $\text{H} \rightarrow \text{L}$ (decreasing similarity), $\text{L} \rightarrow \text{H}$ (increasing similarity), $\text{L} \rightarrow \text{L}$ (consistently low similarity), and $\text{H} \rightarrow \text{H}$ (consistently high similarity).
As shown in Figure~\ref{fig_four_subsets}, only 35.9\% of negatives follow the desired $\text{H} \rightarrow \text{L}$ pattern, suggesting that a limited portion of negatives are consistently pushed apart. The remaining negatives do not exhibit a decreasing trend: 14.5\% remain consistently low, 31.9\% remain consistently high, and 17.8\% unexpectedly increase over training. 
Representative cases in $\text{H} \rightarrow \text{H}$ and $\text{L} \rightarrow \text{H}$ subsets are shown in Figure~\ref{fig:similarity_case}, and detailed classification criteria are provided in Appendix~\ref{sec_classification}.

Overall, these observations suggest that negative pairs contribute unevenly to learning, and prioritizing informative negatives with high contrastive utility is crucial for improving cross-modal alignment. 
To this end, we introduce a Pair Selection strategy that systematically filters candidate negatives with high contrastive utility.

\section{Method}\label{sec_method}

% % 如Figure ~\ref{fig:pipeline},在完成Step 1后(对应Section ~\ref{sec:pre expirement})，我们在本节中，我们首先在Section ~\ref{subsec:framework}介绍我们的SLT Model的结构，之后在Section ~\ref{subsec:PS}中介绍Step 2中的样本选择策略。
% As illustrated in Figure~\ref{fig:pipeline}, following the completion of Step 1 (detailed in Section~\ref{sec:pre-experiment}), this section presents the core components of our proposed method. We first elaborate on the SCL strategy employed in Step 2, as described in Section~\ref{subsec:PS}. Subsequently, introduce the architecture of our SLT model in Section~\ref{subsec:framework}.

% As illustrated in Figure~\ref{fig:pipeline}, we first train a Contrastive Learning Model on Video-Text data (Step 1 in Section~\ref{sec:CLmodel}), and then a negative Pair Selection is performed with the proposed strategy (Step 2 in Section~\ref{subsec:PS}). Finally, we elaborate on the Selective Contrastive Learning for SLT (Step 3 in Section~\ref{subsec:framework} and Section~\ref{sec_slt_ft}).

As illustrated in Figure~\ref{fig:pipeline}, we first train a reference video-text contrastive model (Step 1, in Section~\ref{sec_preliminary}). We then apply the proposed Pair Selection strategy to construct mini-batches with informative and semantically valid negatives (Step 2, in Section~\ref{subsec:PS}). Finally, we present the selective contrastive training framework and its fine-tuning in SLT (Step 3, in Section~\ref{subsec:framework}).

\subsection{Pair Selection}\label{subsec:PS}

\begin{figure*}[t]
\centering
\includegraphics[width=\textwidth]{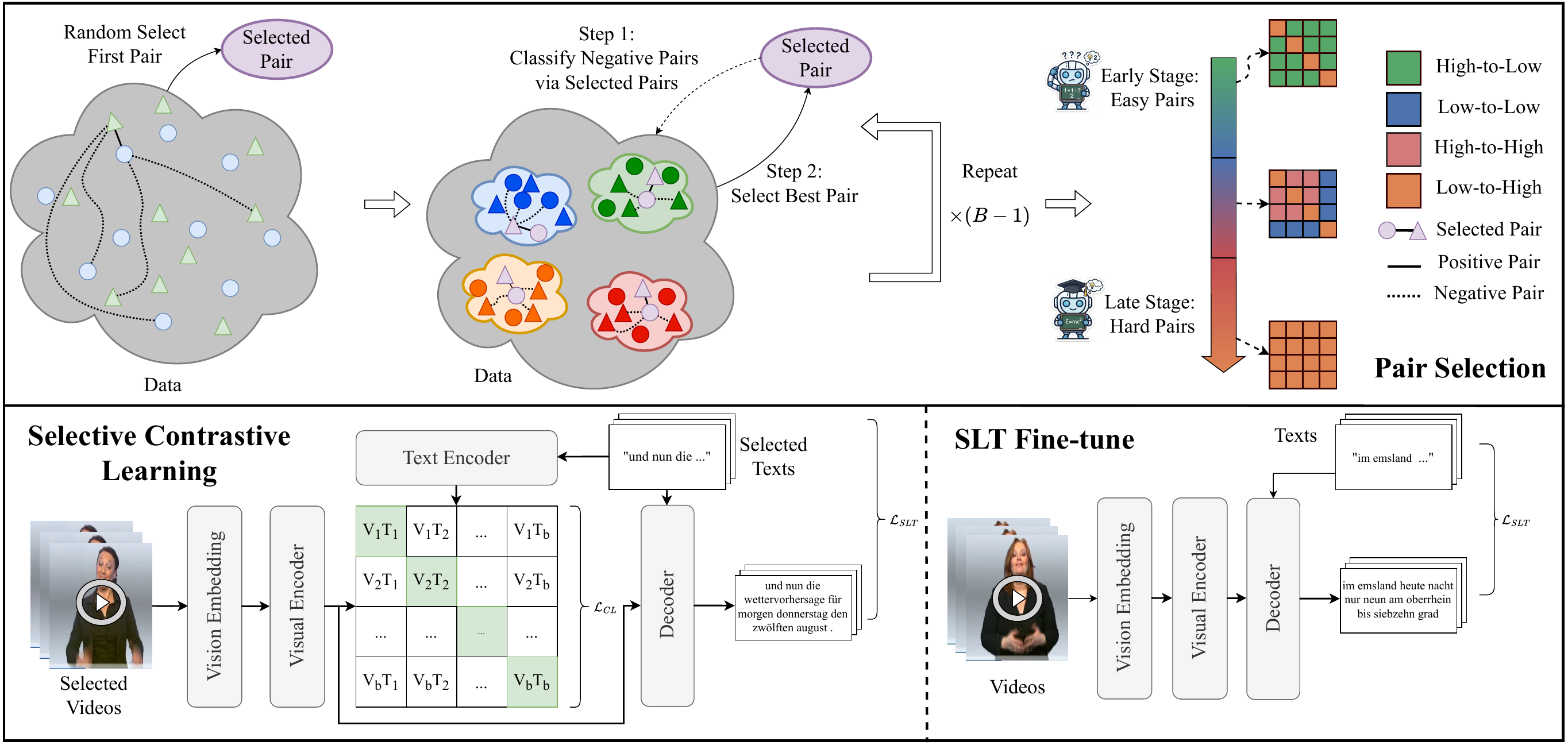}  % 使用 \textwidth
\caption{\textbf{Overview of the proposed SCL-SLT framework.} \textbf{(Top)} Illustration of the Pair Selection strategy. During batch construction, we initialize with a random positive pair and iteratively select subsequent pairs by evaluating candidates against the current selection. The process follows a curriculum learning~\cite{bengio2009curriculum} schedule, progressively transitioning from ``Easy Pairs'' in early stages to ``Hard Pairs'' in later stages. \textbf{(Bottom)} In the first stage, \textbf{Selective Contrastive Learning (SCL)} utilizes selected pairs to align visual and textual modalities. In the second stage, \textbf{SLT Fine-tuning}, the auxiliary Text Encoder and alignment module are detached, allowing the model to focus exclusively on optimizing translation performance.}
\label{fig:train_framework}
\end{figure*}

Inspired by the preliminary findings, we propose a Pair Selection mechanism that improves alignment efficiency by prioritizing informative negatives while mitigating the impact of noisy or semantically invalid negatives.
We adopt a curriculum strategy~\cite{bengio2009curriculum} that progressively shifts the selected negatives from \emph{easy} pairs to \emph{hard} pairs, corresponding to increasingly challenging similarity dynamics (Figure~\ref{fig:train_framework}, top).

\paragraph{Trajectory-based difficulty proxy.}
We quantify negative-pair difficulty using the change in similarity over training.
Let $\delta_{i,j}$ denote the \emph{approximate} similarity change:
\begin{equation}\label{eq:dis}
  \delta_{i,j} = \hat{s}^{K}(V_i,T_j)-\hat{s}^{0}(V_i,T_j),
\end{equation}
where $\hat{s}^{k}(V_i,T_j)$ denotes the \emph{approximate} similarity between the $i$-th sign video and the $j$-th text computed with the $k$-th checkpoint of the reference contrastive model (Step~1 in Figure~\ref{fig:pipeline}), and $K$ denotes the final checkpoint.
We define the batch score as the aggregate summation over all negative pairs within a batch:
\begin{equation}
\label{eq:batch_score}
\mathrm{Score}_{\mathcal{C}} = \sum_{i=1}^{B} \sum_{\substack{j=1, j \neq i}}^{B} \delta_{i, j},
\end{equation}
and aim to construct mini-batches whose scores follow a curriculum ratio $\alpha \in[0,1]$.
We adopt a linear schedule $\alpha = e/E$, where $e$ denotes the current epoch and $E$ denotes the total number of epochs in the selective contrastive training stage.
A formal definition of $\delta_{i,j}$ is provided in Appendix~\ref{sec_classification}.

\begin{algorithm}[t]
\caption{Curriculum-based Pair Selection}
\label{alg:pair_selection}
\begin{algorithmic}[1]
\State \textbf{Input:} dataset $\mathcal{D}=\{(V_i,T_i)\}_{i=1}^{N}$, curriculum ratio $\alpha\in[0,1]$, batch size $B$
\State \textbf{Output:} batch list $\mathbb{B}$
\State $\mathcal{D}'\gets \mathcal{D}$; $\mathbb{B}\gets \emptyset$
\While{$\mathcal{D}'\neq \emptyset$}
    \State sample $s_0\sim \mathrm{Uniform}(\mathcal{D}')$
    \State $\mathcal{C}\gets \{s_0\}$; $\mathcal{D}'\gets \mathcal{D}'\setminus\{s_0\}$
    \While{$|\mathcal{C}|<B$ \textbf{and} $\mathcal{D}'\neq\emptyset$}
        \State $s\gets \textsc{SelectByScore}(\mathcal{D}',\mathcal{C},\alpha)$
        \State $\mathcal{C}\gets \mathcal{C}\cup\{s\}$; $\mathcal{D}'\gets \mathcal{D}'\setminus\{s\}$
    \EndWhile
    \State $\mathbb{B}\gets \mathbb{B}\cup\{\mathcal{C}\}$
\EndWhile
\State \textbf{return} $\mathbb{B}$
\end{algorithmic}
\end{algorithm}

\paragraph{Relation to trajectory categories.}
Our curriculum is driven by the signed similarity change $\delta_{i,j}$ (Equation~\ref{eq:dis}), which implicitly orders negative pairs by their trajectory patterns.
Following the taxonomy in Section~\ref{sec_preliminary}, we refer to \emph{easy} negatives as those that are either pushed away over training (H~$\rightarrow$~L, decreasing similarity) or remain consistently dissimilar (L~$\rightarrow$~L, consistently low similarity), and \emph{hard} negatives as those that stay highly similar (H~$\rightarrow$~H, consistently high similarity) or become more similar over training (L~$\rightarrow$~H, increasing similarity). 
Since $\delta_{i,j}$ can be negative for L~$\rightarrow$~H pairs and positive for H~$\rightarrow$~L pairs, selecting negatives by increasing score naturally yields a curriculum that starts from decreasing-similarity negatives and gradually shifts towards increasing-similarity negatives as $\alpha$ increases.
Pairs in the L~$\rightarrow$~L and H~$\rightarrow$~H regimes typically exhibit small $|\delta_{i,j}|$ magnitudes; thus, while they are not explicitly excluded by our selection rule, they tend to be less influential under a trajectory-change-driven curriculum.

\paragraph{Incremental score for greedy batch construction.}
Given a training dataset $\mathcal{D}=\{(V_i,T_i)\}_{i=1}^{N}$, exhaustively searching over all possible batch constructions to optimize the global score is computationally intractable due to combinatorial explosion and memory overhead, especially when $B\ll N$.
We therefore cast negative selection as a batch construction problem and adopt a greedy procedure (Algorithm~\ref{alg:pair_selection}).
Specifically, given a partially constructed batch $\mathcal{C}$ and a candidate pair $s_u=(V_u,T_u)\in\mathcal{D}'$, we define its \emph{incremental score} as the score increase incurred by adding $s_u$ into $\mathcal{C}$:
\begin{equation}\label{eq:inc_score}
\begin{aligned}
\Delta(s_u;\mathcal{C}) 
&= \mathrm{Score}_{\mathcal{C}\cup\{s_u\}} - \mathrm{Score}_{\mathcal{C}} \\
&= \sum_{(V_i,T_i)\in \mathcal{C}}\left(\delta_{i,u}+\delta_{u,i}\right),
\end{aligned}
\end{equation}
where $\delta_{i,u}$ measures the approximate similarity change between $V_i$ and $T_u$, and $\delta_{u,i}$ is defined analogously between $V_u$ and $T_i$ (Equation~\ref{eq:dis}).
Accordingly, $\textsc{SelectByScore}(\mathcal{D}',\mathcal{C},\alpha)$ computes $\Delta(s_u;\mathcal{C})$ for all $s_u\in\mathcal{D}'$, sorts candidates by $\Delta$ in ascending order, and selects the candidate at the $\alpha$-quantile rank, i.e., $r=\lfloor \alpha\cdot(|\mathcal{D}'|-1)\rfloor$.
Concretely, we seed each batch with one randomly sampled pair from $\mathcal{D}'$, and then iteratively add the selected candidate until the batch reaches the target size $B$.

\subsection{Selective Contrastive Learning}\label{subsec:framework}
% Our SCL-SLT framework comprises four key components: a Sign Embedding module, a Visual Encoder, a Text Encoder, and a Decoder, as shown in Figure ~\ref{fig:train_framework}. 
% Firstly, a CLIP-like contrastive learning is performed based on the proposed Pair Selection strategy, aiming to bridge the modality gap in SLT. 
% Then, an end-to-end fine-tuning on SLT is subsequently conducted, allowing the model to focus exclusively on optimizing the translation objective where the Text Encoder and the alignment module are detached.
Our SCL-SLT framework consists of four components: a Sign Embedding module, a Visual Encoder, a Text Encoder, and a Decoder, as illustrated in Figure ~\ref{fig:train_framework}. 
% First, a CLIP-like contrastive learning phase is implemented using the proposed Pair Selection strategy, designed to bridge the modality gap in SLT. Subsequently, an end-to-end fine-tuning stage is performed on SLT, enabling the model to focus exclusively on optimizing the translation objective, during which the Text Encoder and the alignment module are decoupled.
We first perform selective contrastive training with PS to learn better-aligned video–text representations, while jointly optimizing the translation objective to preserve generation capability. We then conduct task-specific SLT fine-tuning, where the text branch used for alignment is detached and the model is trained with the translation loss only.

\paragraph{Sign Embedding.}
The Sign Embedding module converts an input sign video clip into a temporal feature sequence.
It comprises a ResNet~\cite{resnet} backbone for spatial feature extraction, followed by two temporal blocks, each consisting of stacked 1D Convolution, Batch Normalization, and ReLU activation (Conv1D/BN/ReLU).

\paragraph{Visiual \& Textual Encoders.}
We adopt a dual-encoder design for video-text representation learning.
The Visual Encoder is a Transformer encoder initialized from a pre-trained sequence-to-sequence language model mBART-large-50~\cite{mbart-cc25} and adapted with LoRA~\cite{LoRA}, enabling it to extract high-level visual representations from the Sign Embedding. 
The Text Encoder, also a Transformer encoder initialized from the same mBART-large-50 module, is kept frozen to provide robust linguistic priors for video-text alignment during the contrastive learning stage.

\paragraph{Video--Text Alignment.}
We adopt a CLIP-style contrastive objective to align the visual and textual feature spaces by increasing similarity for matched pairs and decreasing it for mismatched pairs. For sequence-level alignment, we consider three aggregation strategies to obtain global representations: (1) CLS pooling~\cite{gloss-free-VLP}, (2) mean pooling over the feature sequences~\cite{gloss-free-MMSLT, gloss-free-llavaslt}, and (3) the fine-grained CiCo aggregation protocol~\cite{gloss-free-cico}. A comparative analysis is provided in Section~\ref{par:sim compute}. In our final implementation, the similarity matrices $S_{V2T}$ and $S_{T2V}$ are computed following CiCo, and the contrastive loss is computed as:
\begin{equation}
\begin{split}
\mathcal{L}_{\mathrm{CL}}&=-\frac{1}{2B}(\sum_{i=1}^{B}\log\frac{\exp(S_{V2T}^{i,i}/\tau)}{\sum_{k=1}^{B}\exp(S_{V2T}^{i,k}/\tau)}\\&+\sum_{i=1}^{B}\log\frac{\exp(S_{T2V}^{i,i}/\tau)}{\sum_{k=1}^{B}\exp(S_{T2V}^{i,k}/\tau)}),
\end{split}
\end{equation}
where $B$ is the batch size and $\tau$ is a trainable temperature parameter. 
% During selective contrastive training, we optimize a joint objective $\mathcal{L}_{\mathrm{SCL}}=\mathcal{L}_{\mathrm{SLT}}+\lambda\,\mathcal{L}_{\mathrm{CL}}$, with $\lambda$ controlling the strength of alignment supervision.
% During selective contrastive training, we primarily optimize the contrastive alignment loss $\mathcal{L}_{\mathrm{CL}}$ while retaining the standard translation training signal to maintain generation capability, leading to aligned representations. We set the weight of the auxiliary translation objective to 1.0 in all experiments.
During selective contrastive training, we primarily optimize the contrastive alignment loss $\mathcal{L}_{\mathrm{CL}}$. In addition, we retain the standard translation training signal and set its weight to 1.0 to maintain generation capability while learning aligned representations in all experiments.

\begin{table*}[t]
\centering
% 使用 \small 字号，使表格内容更紧凑
\small 
% 稍微减小列与列之间的默认间距
\setlength{\tabcolsep}{2.5pt} 
% 【修改】：删除了 e2e 的 c 列，总列数变为 11 列
% 结构：Method(l) + 5列(Phoenix) + 间隔 + 5列(CSL)
\begin{tabularx}{\textwidth}{l YYYYY @{\hspace{1em}} YYYYY}
\toprule
% 【修改】：调整表头，移除 e2e，调整 cmidrule 跨度 (2-6 和 7-11)
\multirow{2}{*}{\textbf{Method}} & \multicolumn{5}{c}{\textbf{PHOENIX14T}} & \multicolumn{5}{c}{\textbf{CSL-Daily}} \\ 
\cmidrule(lr){2-6} \cmidrule(lr){7-11}

% 【修改】：将 R 移动到 B1 前面
 & \textbf{R} & \textbf{B1} & \textbf{B2} & \textbf{B3} & \textbf{B4} & \textbf{R} & \textbf{B1} & \textbf{B2} & \textbf{B3} & \textbf{B4} \\ 
% \midrule
\noalign{\vskip 1pt} \hline \noalign{\vskip 2pt}
% 【修改】：multicolumn 跨度改为 11
% \rowcolor{gray!20} \multicolumn{11}{c}{GFSLT \textit{w/o} VLP  } \\ 
\multicolumn{11}{c}{\cellcolor[HTML]{EFEFEF} \textit{GFSLT} \textit{w/o} VLP  } \\ 
% \midrule
\noalign{\vskip 2pt} \hline \noalign{\vskip 2pt}
NSLT~\citeyearpar{PHOENIX14T} & 29.70 & 27.10 & 15.61 & 10.82 & 8.35  & - & - & - & - & - \\
NSLT+Bahdanau~\citeyearpar{PHOENIX14T,NMT} & 31.80 & 32.24 & 19.03 & 12.83 & 9.58 & - & - & - & - & - \\
NSLT+Luong~\citeyearpar{PHOENIX14T,gloss-free-NSLT+Luong} & 30.70 & 29.86 & 17.52 & 11.96 & 9.00 & 34.54 & 34.16 & 19.57 & 11.84 & 7.56 \\
SLRT$^\star$~\citeyearpar{gloss-SLT-SLRT} & 31.10 & 30.88 & 18.57 & 13.12 & 10.19 & 19.67 & 20.00 & 9.11 & 4.93 & 3.03 \\ 
MMTLB$^\dagger$~\citeyearpar{gloss-SLT-MMTLB} & 38.60 & 40.57 & 26.99 & 19.58 & 15.18 & 26.70 & 27.22 & 15.90 & 10.61 & 7.61 \\
GASLT~\citeyearpar{gloss-free-GASLT} & 30.86 & 39.07 & 26.74 & 21.86 & 15.74 & 20.35 & 19.90 & 9.94 & 5.98 & 4.07 \\
GFSLT~\citeyearpar{gloss-free-VLP} & 40.93 & 41.39 & 31.00 & 24.20 & 19.66 & 35.16 & 37.69 & 23.28 & 14.93 & 9.88 \\
Sign2GPT~\citeyearpar{gloss-free-Sign2GPT} & 48.90 & 49.54 & 35.96 & 28.83 & 22.52 & 42.36 & 41.75 & 28.73 & 20.60 & 15.40 \\
FLa-LLM~\citeyearpar{gloss-free-FLA} & 45.27 & 46.29 & 35.33 & 28.03 & 23.09 & 37.25 & 37.13 & 25.12 & 18.38 & 14.20 \\
SignLLM~\citeyearpar{gloss-free-SignLLM} & 44.49 & 45.21 & 34.78 & 28.05 & 23.40 & 39.91 & 39.55 & 28.13 & 20.07 & 15.75 \\
\hdashline
\noalign{\vskip 2pt}
\textbf{SCL-SLT(Ours)} & 46.33 & 48.00 & 37.36 & 30.23 & 25.30 & 48.53 & 50.36 & 36.88 & 27.69 & 21.41 \\
% \midrule
\noalign{\vskip 1pt} \hline \noalign{\vskip 2pt}
% \rowcolor{gray!20} \multicolumn{11}{c}{\textit{GFSLT} \textit{w/} VLP} \\
\multicolumn{11}{c}{\cellcolor[HTML]{EFEFEF} \textit{GFSLT} \textit{w/} VLP  } \\ 
\noalign{\vskip 2pt} \hline \noalign{\vskip 2pt}
% \midrule
GFSLT-VLP~\citeyearpar{gloss-free-VLP} & 42.49 & 43.71 & 33.18 & 26.11 & 21.44 & 36.44 & 39.37 & 24.93 & 16.26 & 11.00 \\
GFSLT-VLP-SignCL~\citeyearpar{gloss-free-signcl} & 49.04 & 49.76 & 36.85 & 29.97 & 22.74 & 48.92 & 47.47 & 32.53 & 22.62 & 16.16 \\
% 【修改】：只保留textbf，去掉underline，R移到最前
LLAVA-SLT~\citeyearpar{gloss-free-llavaslt} & 50.44 & 51.20 & 37.51 & 29.39 & 23.43 & \textbf{51.26} & 52.15 & 36.24 & 26.47 & 20.42 \\
C$^2$RL~\citeyearpar{gloss-free-c2rl} &\textbf{50.96} & \textbf{52.81} & \textbf{40.20} & \textbf{32.20} & \textbf{26.75} & 48.21 & 49.32 & 36.28 & 27.54 & 21.61 \\
SpaMo~\citeyearpar{gloss-free-Spamo}  & 46.57 &  49.80 & 37.32 & 29.50 & 24.32 & 47.46&  48.90 & 36.90 & 26.78 & 20.55  \\
MMSLT~\citeyearpar{gloss-free-MMSLT} & 47.97 & 48.92 & 38.12 & 30.79 & 25.73 & 48.92 & 49.87 & 36.37 & 27.29 & 21.11 \\
% \midrule
\hdashline
\noalign{\vskip 2pt}
% \textbf{CL-SLT(Ours)} & 43.55 & 44.74 & 33.91 & 26.84 & 22.03 & 47.77 & 49.37 & 35.96 & 26.90 & 20.59 \\
% \textbf{CL-SLT-F(Ours)} & 46.13 & 47.40 & 37.09 & 29.92 & 25.01 & 48.34 & 49.67 & 36.32 & 27.10 & 20.70 \\
% \hdashline
% \noalign{\vskip 2pt}
% \textbf{SCL-SLT(Ours)} & 46.31 & 48.36 & 37.27 & 29.76 & 24.59 & 48.53 & 50.36 & 36.88 & 27.69 & 21.41 \\
% % 【修改】：只保留textbf，去掉underline，R移到最前
% \textbf{CL-SLT(Ours)} & 46.13 & 47.40 & 37.09 & 29.92 & 25.01 & 48.34 & 49.67 & 36.32 & 27.10 & 20.70 \\
\textbf{SCL-SLT(Ours)} & 47.02 & 48.72 & 38.19 & 31.04 & 26.00 & 51.08 & \textbf{52.81} & \textbf{39.28} & \textbf{29.82} & \textbf{23.25} \\

\bottomrule
\end{tabularx}
\caption{Performance comparison with state-of-the-art gloss-free SLT methods on the PHOENIX14T and CSL-Daily sets. R and B$n$ denote ROUGE and BLEU-$n$, respectively. $^\dagger$ indicates our reproduction under the gloss-free setting. For SLRT$^\star$, results are sourced from~\citet{gloss-free-GASLT} for PHOENIX14T and~\citet{gloss-free-VLP} for CSL-Daily. The best results are highlighted in \textbf{bold}.}
\label{main_results}
\end{table*}

% \subsection{SLT Fine-tuning}\label{sec_slt_ft}
\paragraph{SLT Fine-tuning.}\label{sec_slt_ft}
The Decoder is a LoRA-adapted pre-trained seq2seq decoder.
Conditioned on the sign video $V$ and previously generated tokens $y_{<t}$, it generates the target sentence autoregressively.
In the SLT fine-tuning stage, we detach the text branch used for alignment and optimize the translation objective only:
\begin{equation}
\label{eq:slt}
\mathcal{L}_{\mathrm{SLT}} = -\sum_{t=1}^{|y|}\log P(y_t \mid V, y_{<t}).
\end{equation}

% ========================= SCL end =========================

\section{Experiments}
\label{sec_experiments}
\subsection{Datasets and Evaluation Metrics}
% \paragraph{Datasets.} We evaluate our method on two widely used benchmarks: RWTH-PHOENIX-Weather-2014T (PHOENIX14T)~\cite{PHOENIX14T} and CSL-Daily~\cite{gloss-SLT-BN-TIN-Transf}. \textbf{PHOENIX14T} focuses on German Sign Language within the weather forecast domain. It contains 7,096/519/642 pairs (Train/Dev/Test) with a vocabulary of 2,887. \textbf{CSL-Daily} is a large-scale Chinese Sign Language dataset covering daily communication topics. Characterized by its diverse multi-signer variability, it comprises 18,401/1,077/1,076 samples with a vocabulary size of 2,343.
\paragraph{Datasets.} We evaluate our method on two benchmarks: \textbf{PHOENIX14T}~\cite{PHOENIX14T}, a German Sign Language dataset in the weather domain, and \textbf{CSL-Daily}~\cite{gloss-SLT-BN-TIN-Transf}, a large-scale Chinese dataset covering daily topics. PHOENIX14T contains 7,096/519/642 pairs (Train/Dev/Test) with a vocabulary of 2,887. CSL-Daily, noted for its multi-signer diversity, comprises 18,401/1,077/1,076 samples with a vocabulary size of 2,343.

\paragraph{Evaluation Metrics.} Following previous studies~\cite{gloss-free-MMSLT,gloss-free-VLP}, we use BLEU~\cite{BLEU} and ROUGE~\cite{ROUGE} to measure the translation performance.

% \noindent \textbf{Implementation Details.}视频数据处理，首先对视频数据进行了下采样，采样率为$k$，则将$T$帧的视频划分为$k\times T$个clip，在训练时，在每个clip中随机选取一帧，在推理时选取每个clip的第一帧，所有帧都首先变形为$256\times 256$，然后使用随机/中心裁剪为$224\times 224$。我们使用在ImageNet~\cite{imagenet}上预训练的Resnet18~\cite{resnet}，与时序模块Conv1d-BN-ReLU layers. Conv1d 的卷积核大小为3步长为1。Vision Encoder、Text Encoder与Text Decoder都为12层，隐藏层维度为1024，前馈层维度为4096，有着16个attention head，其初始化权重来自于Mbart-large-50-many-to-many-mmt。\footnote{https://huggingface.co/facebook/mbart-large-50-many-to-many-mmt.}
\subsection{Implementation Details}
\paragraph{Data Processing.} We apply $4\times$ temporal downsampling, selecting a random frame per clip during training and the first frame during inference. Spatially, frames are resized to $256 \times 256$ and then cropped to $224 \times 224$, using random cropping for training and center cropping for inference.

% \paragraph{SLT Model.}
% The Sign Embedding module utilizes a ResNet-18~\cite{resnet} (pre-trained on ImageNet~\cite{imagenet}) followed by a temporal block comprising Conv1D (kernel=3, stride=1), BatchNorm, and ReLU layers.
% The Text Encoder is instantiated with the pre-trained mBART encoder~\cite{mbart-m2m}, which remains frozen throughout training.
% Regarding the Translation Backbone, we investigate two distinct configurations:
% (1) A Vanilla Transformer consisting of a 3-layer encoder and decoder ($d_{model}=512, d_{ffn}=2048$, 8 heads); and
% (2) A LoRA-tuned LLM initialized from \texttt{mbart-large-50}\footnote{\url{https://huggingface.co/facebook/mbart-large-50-many-to-many-mmt}}. This model features a 12-layer encoder and decoder ($d_{model}=1024, d_{ffn}=4096$, 16 heads), adapted via LoRA with rank $r=16$ and $\alpha=32$.
% \paragraph{Model.} We use a ResNet-18~\cite{resnet} pretrained on ImageNet~\cite{imagenet} followed by a temporal block (Conv1D-BN-Relu-Maxpooling),  for sign embedding. The Visual Encoder, Text Encoder and Decoder is all initialized with \texttt{mbart-large-50}~\cite{mbart-m2m} ($L=12, d_{model}=1024, d_{ffn}=4096$, 16 heads). Specifically, the Text Encoder is frozen, while the Visual Encoder and Decoder are fine-tuned via LoRA ($r=16, \alpha=32$).
\paragraph{Model Architecture.} For sign embedding, we employ a ResNet-18~\cite{resnet} pretrained on ImageNet~\cite{imagenet}, followed by a temporal block (Conv1D-BN-Relu). The Visual Encoder, Text Encoder, and Decoder are initialized with the pre-trained \texttt{mbart-large-50}~\cite{mbart-m2m} ($L=12, d_{model}=1024, d_{ffn}=4096$, 16 heads). During training, the Text Encoder is frozen, while the Visual Encoder and Decoder are fine-tuned via Low-Rank Adaptation (LoRA) with $r=16$ and $\alpha=32$.

% \paragraph{Hyperparameters.} We optimize both stages using AdamW~\cite{Adamw} with a batch size of 16 and a label smoothing factor of 0.2. The learning rate is initialized at $1 \times 10^{-4}$ and decays following a cosine schedule~\cite{cosine}. The training duration is set to 80 epochs for Stage 1 and 200 epochs for Stage 2. During inference, we utilize a beam size of 8.
\paragraph{Hyperparameters.} We optimize both stages using AdamW~\cite{Adamw} with 0.2 label smoothing, and batch sizes of 16 and 8 for Stages 1 and 2. The initial learning rate is $1 \times 10^{-4}$ with a cosine decay~\cite{cosine}. The training duration is set to 80 epochs for Stage 1 and 200 epochs for Stage 2. During inference, we utilize a beam size of 8.

\subsection{Main Results}

Table~\ref{main_results} presents a quantitative comparison with state-of-the-art gloss-free SLT methods. Our SCL-SLT demonstrates strong overall performance, establishing new state-of-the-art records on the CSL-Daily dataset and achieving highly competitive results on PHOENIX14T.

\paragraph{Results on PHOENIX14T.} 
SCL-SLT demonstrates highly competitive performance on the PHOENIX14T benchmark. Under the \textit{w/o} VLP setting, it establishes a new state-of-the-art with a BLEU-4 of \textbf{25.30}. With VLP, SCL-SLT reaches \textbf{26.00}. While C$^2$RL~\cite{gloss-free-c2rl} achieves 26.75 via auxiliary tasks, our approach yields comparable results through a fundamentally streamlined paradigm: maximizing intrinsic data utility via refined negative sampling. Furthermore, SCL-SLT surpasses the strong LLM-centric LLAVA-SLT~\cite{gloss-free-llavaslt} by \textbf{2.57} in BLEU-4 (26.00 vs. 23.43).

% \paragraph{Results on PHOENIX14T.}
% SCL-SLT achieves the highest score in the most critical metric, reaching a BLEU-4 of \textbf{25.98}. This result surpasses the previous best method, MMSLT~\cite{gloss-free-MMSLT} (25.73), and significantly outperforms VLP-Free methods such as SignLLM~\cite{gloss-free-SignLLM} (23.40). Comparisons with the recent strong competitor LLAVA-SLT~\cite{gloss-free-llavaslt} reveal distinct performance characteristics: while LLAVA-SLT exhibits higher scores in ROUGE and lower-order n-grams (BLEU-1), our method demonstrates a significant advantage in higher-order metrics, outperforming it by \textbf{2.55} in BLEU-4 (25.98 vs. 23.43). 
%This discrepancy suggests that our SCL strategy is more effective at modeling long-range dependencies and generating syntactically coherent long sentences (high B4).

\paragraph{Results on CSL-Daily.}
On the more challenging CSL-Daily benchmark, our method achieves a remarkable \textbf{23.25 BLEU-4} with a substantial margin of \textbf{2.14} over the previous best C$^2$RL (21.61). 
While the LLM-based method LLAVA-SLT remains competitive in ROUGE (51.26 vs. 51.08), our approach significantly outperforms it in all precision-oriented BLEU metrics (e.g., +2.83 in BLEU-4).

% \paragraph{Superiority over VLP-Free.}
%只将SCL作为端到端手语翻译的辅助手段，SCL-SLT也在VLP-Free中表现亮眼，与多阶段的Sign2GPT~\cite{gloss-free-Sign2GPT}、FLa-LLM~\cite{gloss-free-FLA}、SignLLM ~\cite{gloss-free-SignLLM}相比，
\paragraph{Superiority over Methods \textit{w/o} VLP.} SCL-SLT achieves state-of-the-art performance on both PHOENIX14T and CSL-Daily benchmarks under the \textit{w/o} VLP setting. Notably, SCL-SLT achieves a BLEU-4 score of \textbf{25.30} on PHOENIX14T and \textbf{21.41} on CSL-Daily, surpassing the strongest competitor (SignLLM) by substantial margins of \textbf{1.90} and \textbf{5.66}, respectively.

% \paragraph{Superiority over Baselines \textit{w/} VLP.}
% % A critical advantage of SCL-SLT lies in its efficiency and reliance on intrinsic data signals. Compared to other VLP-based methods, our approach achieves superior performance without relying on external auxiliary information. For instance, MMSLT necessitates additional frame-level motion descriptions to guide generation. In contrast, SCL-SLT relies purely on a refined negative sampling strategy to mine informative pairs from the dataset itself. Our empirical results demonstrate that by focusing solely on robust cross-modal alignment, SCL-SLT achieves comparable or even superior performance without the need for introducing extra information.
% A critical advantage of SCL-SLT is its reliance on intrinsic data signals rather than external auxiliary information. Unlike methods such as MMSLT and C$^2$RL, which necessitates additional frame-level motion descriptions, SCL-SLT employs a refined negative sampling strategy to mine informative pairs directly from the dataset. Our results demonstrate that focusing on robust cross-modal alignment is sufficient to achieve superior performance without introducing extra complexity.

\paragraph{Superiority over Baselines \textit{w/} VLP.}
A critical advantage of SCL-SLT lies in its elegant reliance on intrinsic data signals. While recent strong baselines depend on external annotations (e.g., frame-level motion descriptions in MMSLT~\cite{gloss-free-MMSLT}) or complex auxiliary objectives (e.g., Explicit Context Learning in C$^2$RL~\cite{gloss-free-c2rl}), SCL-SLT avoids such architectural overhead. Instead, it employs a refined negative sampling strategy to systematically mine highly informative pairs directly from the existing dataset. Our empirical results demonstrate that maximizing data utility to establish robust cross-modal alignment is sufficient to achieve superior performance, entirely eliminating the need for extraneous complexity.

\begin{table}[t]
\centering
\small % 保持小字号，这很好

% 1. 使用 tabularx，设置总宽为 \columnwidth
\begin{tabularx}{\columnwidth}{l Y Y Y Y} 
\toprule

\multirow{2}{*}{\textbf{Method}} & \multicolumn{2}{c}{\textbf{PHOENIX14T}} & \multicolumn{2}{c}{\textbf{CSL-Daily}}\\
\cmidrule(lr){2-3}\cmidrule(lr){4-5}
& \textbf{R} & \textbf{B4} & \textbf{R} & \textbf{B4} \\
\midrule
% 2. 记得给 [cls] 加上花括号防止超时
BaseLine(End-to-End)    & 41.81 & 21.97 & 41.04 & 16.31 \\
~\textit{w/} CL  & 43.55 & 22.03 & 47.77 & 20.59\\
~\textit{w/} SCL  & 46.33 & 25.30 & 48.53 & 21.41\\
\midrule
CL-SLT & 46.13 & 25.01 & 48.34 & 20.70 \\
SCL-SLT (Ours)    & 47.02 & 26.00 & 51.08 & 23.25 \\
\bottomrule
\end{tabularx}
\caption{Ablation study on the effectiveness of the SCL strategy. (Top) End-to-End SLT. (Bottom) Vision-Language Pretraining SLT.
% The results show that the selective strategy significantly outperforms the random in-batch negative construction.
% \textit{w/}~SCL 
% significantly outperforms \textit{w/}~CL, 
% validating the superiority of mining informative negative samples over random sampling.
}
\label{tab:ab_scl}
\end{table}

\begin{table}[t]
\centering

\small

\label{tab:redundancy_stats}
% 使用 tabular 即可，不需要 tabularx，让列宽自适应内容更美观
\setlength{\tabcolsep}{9pt}
\begin{tabularx}{\columnwidth}{ccccc}
\toprule
\multirow{2}{*}{\textbf{Dataset}} & \multicolumn{4}{c}{
\textbf{Number of Videos for Unique Text}
} \\
\cmidrule(lr){2-5} 
& \textbf{1} & \textbf{2} & \textbf{3} & \textbf{$\geq$ 4} \\
\midrule
PHOENIX14T &6,811& 21 & 5 & 16 \\
CSL-Daily  &45& 1,480 & 4,850 & 203 \\
\bottomrule
\end{tabularx}
\caption{\textbf{Statistics of target redundancy in SLT benchmarks.} We report the number of unique target sentences associated with $N$ different sign videos. Notably, CSL-Daily exhibits a high degree of redundancy, with most texts corresponding to multiple videos (e.g., $N=3$).}
\label{tab:repeat data}
\end{table}

\begin{table}[t]
\centering
\small

\begin{tabularx}{\columnwidth}{lYYYYY}
\toprule
\textbf{Method} & \textbf{R}  &  \textbf{B1} & \textbf{B2} & \textbf{B3} & \textbf{B4} \\
\midrule
Random   & 43.55 & 44.74 & 33.91 & 26.84 & 22.03\\
\midrule
Hard-Only & 28.94 & 31.54 & 21.46 & 15.90 & 12.41 \\
Easy-Only & 44.93  & 46.57 & 35.92 & 28.83 & 24.00 \\
Linear & 46.31 & 48.36 & 37.27 & 29.76 & 24.59 \\
Sqrt & 45.86 & 47.56 & 36.73 & 29.53 & 24.54 \\
Log & 46.33 & 48.00 & 37.36 & 30.23 & 25.30  \\

\bottomrule
\end{tabularx}
\caption{Ablation study on pair selection strategies.}
% \caption{\textbf{Ablation study on pair selection strategies.} We compare our proposed Curriculum strategy against static baselines (Random). The results demonstrate that the Curriculum strategy achieves superior performance, whereas the Hard-Only strategy leads to significant degradation.}
\label{tab:PS-method}
\end{table}

\subsection{Ablation Studies}

In this section, we conduct ablation studies to validate the effectiveness of the proposed method. Unless otherwise specified, all ablation experiments are performed on the PHOENIX14T test set.

% \paragraph{Effectiveness of SCL.}\label{par:effect SCL} Table~\ref{tab:ab_scl} compares our Selective Contrastive Learning (\textit{w/}SCL) against the End-to-End baseline . On PHOENIX14T, \textit{w/}SCL yields a substantial gain of \textbf{2.62 BLEU-4}, falling just 0.42 points short of the fine-tuned CL-SLT, proving the exceptional efficacy of our alignment strategy even without fine-tuning. On CSL-Daily, while the initial gain over \textit{w/}CL is 0.82 BLEU-4, this margin expands significantly to \textbf{2.55 BLEU-4} after fine-tuning (comparing SCL-SLT vs. CL-SLT). This confirms that SCL learns robust representations that generalize better to the final translation task. Further discussion on the impact of contrastive learning is provided in Section~\ref{sec:cl impaction}.

% To explicitly evaluate how different strategies influence contrastive alignment, the following ablation experiments are performed using the \textit{w/} SCL configuration.

\paragraph{Effectiveness of SCL.}\label{par:effect SCL} 
% Table~\ref{tab:ab_scl} compares our Selective Contrastive Learning (\textit{w/}SCL) against the End-to-End baseline. On PHOENIX14T, \textit{w/}SCL yields a substantial gain of \textbf{2.62 BLEU-4}. Notably, this result trails the fully fine-tuned CL-SLT by only 0.42 points, demonstrating the exceptional efficacy of our alignment strategy even prior to fine-tuning. On CSL-Daily, while the initial gain over the random baseline (\textit{w/}CL) is 0.82 BLEU-4, this margin widens significantly to \textbf{2.55 BLEU-4} after fine-tuning (comparing SCL-SLT vs. CL-SLT). This confirms that SCL facilitates the learning of robust representations that generalize better to the final translation task. Further discussion on the impact of contrastive learning is provided in Section~\ref{sec:cl impaction}.
%Table~\ref{tab:ab_scl} compares our Selective Contrastive Learning (\textit{w/}SCL) against the End-to-End baseline. \textit{w/}CL 在PHOENIX14T上的提升仅有0.06，但在CSL-Daily上提升了4.28，这是因为PHOENIX14T集中在天气预报一个领域，其不同样本的差异主要集中在领域类的细粒度差异，而CSL-Daily涵盖了多个话题，每个话题内有一定量的样本，通过话题之间的对比就能够取得很好的效果。\textit{w/}SCL上的表现很好的验证了这一点，在相同领域/话题中，需要对负样本进行更细致的筛选，从而取得更好的效果，相较于\textit{w/}CL，\textit{w/}SCL在PHOENIX14T上提升了2.56，在CSL-Daily上仅提升了0.82，PHOENIX14T领域集中相似样本数量多，而CSL-Daily话题分布广，但话题内样本数量少。在SCL-SLT与CL-SLT的对比中，CSL-Daily上提升了2.55个BLEU4，这表示即使\textit{w/}SCL在端到端的训练中BLEU4提升有限，但其学到了更加鲁棒的特征表示。
Table~\ref{tab:ab_scl} reveals that \textit{w/}~CL works well on the multi-topic CSL-Daily (+4.28) but fails on the weather-focused PHOENIX14T (+0.06), as random sampling lacks the granularity to distinguish similar intra-domain samples. \textit{w/}~SCL resolves this by filtering for informative negatives, boosting PHOENIX14T performance by \textbf{2.56} over \textit{w/}~CL. Although the initial gain on CSL-Daily is modest (+0.82) due to sparse intra-topic samples, the fine-tuned SCL-SLT surpasses CL-SLT by a remarkable \textbf{2.55 BLEU-4}. This confirms that even when immediate end-to-end gains are limited, SCL captures superior structural features that are critical for maximizing final translation quality. 
% As shown in Table~\ref{tab:repeat data}, text-to-video multiplicity poses a significant risk of false negatives. Unlike \textit{w/}~CL, which is prone to sampling these identical pairs as negatives, \textit{w/}~SCL incorporates a filtering mechanism to eliminate such collisions.

As Table~\ref{tab:repeat data} shows, text-to-video multiplicity introduces severe false negatives. Unlike the baseline \textit{w/}~CL, our \textit{w/}~SCL filters these identical-pair collisions. Beyond identical texts, high intra-dataset similarity exacerbates the false negative problem, directly amplifying the gains from our Pair Selection (PS). This is prominent on PHOENIX14T (\textit{w/o}~VLP), where the narrow weather domain yields highly homogeneous texts. Furthermore, even in broader datasets like CSL-Daily (\textit{w/}~VLP), inherent sign language vocabulary limits inevitably introduce structural similarities. By navigating these intrinsic similarities, PS learns robust cross-modal representations across diverse domains.

To further assess the influence of distinct contrastive strategies, subsequent ablation experiments are performed under the \textit{w/}~SCL setting.

 % As show in Table~\ref{tab:repeat data}. To investigate the underlying cause, we analyzed the video-text multiplicity. We observe a distinct disparity: on PHOENIX14T, the vast majority ($95.98\%$) of texts correspond to a unique video. Conversely, on CSL-Daily, over \textbf{99.31\%} of texts are associated with multiple sign videos. In standard contrastive learning, these semantically identical pairs are treated as negatives, inevitably introducing severe noise (i.e., false negatives). Our SCL effectively mitigates this issue by filtering out such informative pairs, explaining the significant performance gap.

% \paragraph{Impact of Pair Selection Strategies.}\label{par:pair impact} To investigate the influence of different curriculum scheduling Parameter $\alpha$ on translation performance, we use distinct settings: Random(BaseLine), Easy-Only, Hard-Only, and our proposed Curriculum Learning. As shown in Table~\ref{tab:PS-method}. Compared to the Random baseline, both the Easy-Only and Curriculum strategies yielded improvements. Conversely, the Hard-Only strategy proved detrimental, leading to a significant performance degradation of 7.62 BLEU-4 points below the baseline. Ultimately, our Curriculum Learning approach achieved the best results, outperforming Random selection by +2.56 and the Easy-Only strategy by +0.59 BLEU-4. These findings further validate that not all negative pairs contribute positively to training.
\paragraph{Impact of Curriculum Scheduling Strategies.}\label{par:pair impact} 
As shown in Table~\ref{tab:PS-method}
To investigate the influence of the curriculum scheduling parameter $\alpha$ on translation performance, we evaluate both static and dynamic settings. Specifically, $\alpha=0$ indicates that only the most easily distinguishable simple samples are selected (Easy-Only), while $\alpha=1$ means that only highly challenging hard samples are selected (Hard-Only). Let $k$ denote the current training epoch and $E_{\mathrm{ref}}$ represent the total number of epochs. In addition to the two extreme static strategies, we focus on comparing the following three dynamic scheduling functions, where $\alpha$ gradually transitions from $0$ to $1$ as training progresses: (1) \textbf{Linear Scheduling (Linear: $\alpha = {k}/{E_{\mathrm{ref}}}$)} (2) \textbf{Logarithmic Scheduling (Log: $\alpha = \ln\left(1 + ({k}/{E_{\mathrm{ref}}}) \cdot (\mathrm{e} - 1)\right)$)} (3) \textbf{Square Root Scheduling (Sqrt: $\alpha = \sqrt{({k}/{E_{\mathrm{ref}}})}$)}.

% \begin{itemize}
%     \item \textbf{Linear Scheduling (Linear):} $\alpha = \frac{k}{E_{\mathrm{ref}}}$. 
%     \item \textbf{Logarithmic Scheduling (Log):} $\alpha = \ln\left(1 + \frac{k}{E_{\mathrm{ref}}} \cdot (\mathrm{e} - 1)\right)$.
%     \item \textbf{Square Root Scheduling (Sqrt):} $\alpha = \sqrt{\frac{k}{E_{\mathrm{ref}}}}$.
% \end{itemize}

\paragraph{Impact of Similarity Computation Methods.}\label{par:sim compute} As show in Table~\ref{tab:sim compute}.To investigate the influence of different similarity matrix calculation strategies, we conducted experiments using three approaches: the \texttt{[CLS]}  Pooling, Mean Pooling, and CiCo~\cite{gloss-free-cico}. As shown in Table~\ref{tab:sim compute}, the CiCo method yields significantly superior performance, outperforming both the \texttt{[CLS]} Pooling and Mean Pooling baselines by a substantial margin.

\begin{table}[t]
\centering
\small % 保持小字号，这很好

% 1. 使用 tabularx，设置总宽为 \columnwidth
\begin{tabularx}{\columnwidth}{l Y Y Y Y Y} 
\toprule
% 注意：这里不需要 resizebox 了
\textbf{Method} & \textbf{R} & \textbf{B1} & \textbf{B2} & \textbf{B3} & \textbf{B4} \\ 
\midrule
% 2. 记得给 [cls] 加上花括号防止超时
Mean Pooling & 28.52  & 30.57 & 20.92 & 15.66 & 12.44\\
\texttt{[CLS]} Pooling & 33.03   & 35.20 & 24.98 & 18.73 & 14.85 \\

CiCo~\citeyearpar{gloss-free-cico}   & 46.33 & 48.00 & 37.36 & 30.23 & 25.30 \\

\bottomrule
\end{tabularx}
% \caption{Ablation study on different Similarity Computation Methods.}
\caption{Ablation study on representation aggregations.}
\label{tab:sim compute}
\end{table}

\begin{table}[t]
\centering
\small
\begin{tabularx}{\columnwidth}{l Y Y Y Y Y} 
\toprule
\textbf{Interval} & \textbf{R} & \textbf{B1} & \textbf{B2} & \textbf{B3} & \textbf{B4} \\ 
\midrule
1  & 45.22 & 47.64 & 36.91 & 29.66 & 24.68 \\
5  & \textbf{46.33} & \textbf{48.00} & \textbf{37.36} & \textbf{30.23} & \textbf{25.30} \\
10 & 15.89 & 16.33 & 10.26 &  7.73 &  6.26 \\
\bottomrule
\end{tabularx}
\caption{Ablation study on the checkpoint sampling interval for trajectory estimation.}
\label{tab:cpk_step}
\end{table}

\paragraph{Impact of Checkpoint Sampling Interval.}
Table~\ref{tab:cpk_step} evaluates the sampling granularity for similarity trajectory estimation. An interval of 5 epochs yields the optimal BLEU-4 of \textbf{25.30}. Sampling too frequently (interval of 1) slightly degrades performance (24.68) by incorporating overly sensitive micro-fluctuations and training noise. Conversely, a sparse interval of 10 causes a catastrophic drop to 6.26 BLEU-4. This indicates that a coarse temporal resolution fails to capture the true dynamics of negative pairs, effectively misguiding the Pair Selection curriculum. Thus, an interval of 5 optimally balances trend capturing and noise smoothing.

\section{Related Work}
\subsection{Gloss-Free SLT with Contrastive Learning} 
GFSLT-VLP~\cite{gloss-free-VLP} first introduced CLIP-based pre-training to SLT, aligning global video features (via \texttt{[CLS]}) with text. CiCo~\cite{gloss-free-cico} improved upon this by performing fine-grained frame-text alignment without the global token constraint. To further boost representation quality, MMSLT~\cite{gloss-free-MMSLT} and C\textsuperscript{2}RL~\cite{gloss-free-c2rl} introduced auxiliary supervision signals, utilizing action descriptions and an ECL loss, respectively. LLaVA-SLT~\cite{gloss-free-llavaslt} combined contrastive alignment with Large Language Models via instruction tuning, achieving robust translation performance.

\subsection{Gloss-Free SLT without Contrastive Learning} To alleviate the dependency on expensive gloss annotations, GASLT~\cite{gloss-free-GASLT} proposes a gloss attention mechanism that implicitly captures gloss-level representations in a fully gloss-free setting. More recently, leveraging the powerful reasoning and generative capabilities of Large Language Models (LLMs), methods such as Sign2GPT~\cite{gloss-free-Sign2GPT} and SignLLM~\citet{gloss-free-SignLLM} have integrated pre-trained LLMs into the SLT framework. These approaches facilitate robust, direct translation from continuous sign videos to spoken sentences by harnessing the linguistic priors of LLMs.

\subsection{Data Selection Strategies}
Recent advancements in NLP highlight that data quality often outweighs quantity~\cite{zhou2023lima}, driving automated selection methods based on model scoring or training dynamics~\cite{chen2023alpagasus, lin2024rho, xia2024less}. Inspired by this paradigm, we extend the focus from sample-level filtering in general NLP to pair-level selection in cross-modal SLT contrastive learning.

\section{Conclusion}
% We propose SCL-SLT, a selective contrastive learning framework for GFSLT. It prioritizes informative negatives while mitigating the impact of noisy or semantically invalid negatives via the Pair Selection mechanism. 
% % 
% Our preliminary analysis revealed the limitations of vanilla contrastive learning with random in-batch negatives. 
% Through a curriculum-based approach to select informative negatives, SCL-SLT effectively suppresses noise and enhances alignment efficiency. 
% Finally, our method achieves state-of-the-art performance on PHOENIX14T and CSL-Daily. 

In this work, we investigate the role of in-batch negatives in CLIP-like contrastive learning in SLT and find that negative pairs exhibit highly uneven training dynamics, leading to unreliable video-text alignment. 
Motivated by this, we proposed SCL-SLT with a curriculum-guided Pair Selection strategy that prioritizes informative negatives while reducing the influence of noisy or semantically invalid ones. Our SCL-SLT improves over the vanilla contrastive learning and achieves new state-of-the-art performance on PHOENIX14T and CSL-Daily. 
% Experiments on PHOENIX14T and CSL-Daily demonstrate consistent gains over vanilla contrastive learning with random in-batch negatives, reaching 24.59/21.41 BLEU under end-to-end training and 25.98/23.25 BLEU with standard vision–language pretraining.

% \section*{Acknowledgment}

\section*{Limitations}
A limitation of our current framework is the reliance on a separately trained Contrastive Learning Model to score and select negative pairs. This additional training stage introduces extra computational overhead and inference latency during the data preparation phase. To mitigate this, future work could explore leveraging powerful off-the-shelf pre-trained language models to compute semantic similarity directly. Replacing the trained reference model with such open-source alternatives would significantly streamline the pipeline and reduce resource consumption.

\section*{Ethics Statement}
Our SCL-SLT framework aims to assist communication for the Deaf and hard-of-hearing community, but it is an experimental exploration of methodology and is not yet ready for real-world production or deployment. Furthermore, as our model is trained on specific regional datasets (PHOENIX14T and CSL-Daily), it may inherit data biases and exhibit degraded performance on out-of-distribution signers, diverse dialects, or unconstrained environments. We exclusively utilized public datasets and advocate for the responsible, privacy-preserving deployment of SLT technologies.

\section*{Acknowledgements}
We are grateful for the efforts and time of the reviewers and the committee. 
This work was supported in part by the National Natural Science Foundation of China under Grant 62476232, Grant 62076211, and in part by First Batch of Projects for the 2025 ``Intergovernmental International Science, Technology and Innovation Cooperation'' of the National Key Research and Development Program of China under Grant 2025YFE0121700.

% \section*{Acknowledgments}

% This document has been adapted
% by Steven Bethard, Ryan Cotterell and Rui Yan
% from the instructions for earlier ACL and NAACL proceedings, including those for
% ACL 2019 by Douwe Kiela and Ivan Vuli\'{c},
% NAACL 2019 by Stephanie Lukin and Alla Roskovskaya,
% ACL 2018 by Shay Cohen, Kevin Gimpel, and Wei Lu,
% NAACL 2018 by Margaret Mitchell and Stephanie Lukin,
% Bib\TeX{} suggestions for (NA)ACL 2017/2018 from Jason Eisner,
% ACL 2017 by Dan Gildea and Min-Yen Kan,
% NAACL 2017 by Margaret Mitchell,
% ACL 2012 by Maggie Li and Michael White,
% ACL 2010 by Jing-Shin Chang and Philipp Koehn,
% ACL 2008 by Johanna D. Moore, Simone Teufel, James Allan, and Sadaoki Furui,
% ACL 2005 by Hwee Tou Ng and Kemal Oflazer,
% ACL 2002 by Eugene Charniak and Dekang Lin,
% and earlier ACL and EACL formats written by several people, including
% John Chen, Henry S. Thompson and Donald Walker.
% Additional elements were taken from the formatting instructions of the \emph{International Joint Conference on Artificial Intelligence} and the \emph{Conference on Computer Vision and Pattern Recognition}.

\bibliography{custom}

\appendix

\section{Classification of Negatives}\label{sec_classification}
To characterize how negative video--text similarities evolve during training, we train a reference contrastive model on PHOENIX14T following the GFSLT-VLP framework~\cite{gloss-free-VLP}.
We save checkpoints every 5 epochs over $E_{\mathrm{ref}}$ training epochs, yielding a checkpoint index set
$\mathcal{K}=\{0,5,\ldots,E_{\mathrm{ref}}\}$, and denote the final checkpoint by $K=\max(\mathcal{K})$.

\paragraph{Similarity trajectories.}
For each checkpoint $k\in\mathcal{K}$, we compute similarities between all video-text pairs and collect negative similarities $\{s^{k}(V_i,T_j)\}_{k\in\mathcal{K}}$ for all $i\in\{1,\ldots,N\}$ and $j\neq i$:
\begin{equation}
\label{eq:cosine_similarity}
s^{k}(V_i, T_j) = \frac{\text{VE}^{k}(V_i) \cdot \text{TE}^{k}(T_j)}{|\text{VE}^{k}(V_i)|_2 |\text{TE}^{k}(T_j)|_2}
\end{equation}
where $\text{VE}^{k}(V_i)$ and $\text{TE}^{k}(T_j)$ denote the high-dimensional global feature vectors extracted by the visual and text encoders for the source video $V_i$ and text $T_j$ at the $k$-th checkpoint, respectively. Here, $\|\cdot\|_2$ represents the $L_2$ norm (i.e., Euclidean length) of the vector, which applies $L_2$ normalization to the feature representations in the denominator. The numerator calculates the dot product of the two vectors. Through this normalization, the ratio is mathematically equivalent to the cosine of the angle between the feature vectors, stably bounding the resulting scalar similarity score within the $[-1, 1]$ interval.

We then fit each negative trajectory using linear least-squares regression:
\begin{equation}
\label{eq:ls_fit}
\min_{a,b}\sum_{k\in\mathcal{K}}\big(s^{k}(V_i,T_j)-(a k+b)\big)^2,
\end{equation}
which yields a smoothed approximation: 
\begin{equation}
\label{eq:approx_sim}
\hat{s}^{k}(V_i,T_j)=a k+b,\quad k\in\mathcal{K}.
\end{equation}

\paragraph{High/Low partition and change magnitude.}
We define the mean final negative similarity as: 
\begin{equation}\label{eq:sim_m}
  \hat{s}_{\mathrm{mean}}
  =\frac{1}{N(N-1)}
  \sum_{i=1}^{N}\sum_{\substack{j=1, j\neq i}}^{N}\hat{s}^{K}(V_i,T_j),
\end{equation}
and label a negative pair as \emph{high} (H) if $\hat{s}^{K}(V_i,T_j)>\hat{s}_{\mathrm{mean}}$ and \emph{low} (L) otherwise.
The similarity change $\delta_{i,j}$ is computed as in Equation~\ref{eq:dis}, and we use a threshold $\epsilon$ to decide whether the change is substantial.
Here, we set $\epsilon=0.2$.

\paragraph{Classification rules.}
Finally, each negative pair is categorized based on its final similarity level and change direction:
\begin{equation}\label{eq:transition_rules}
\begin{cases}
\mathrm{L}\rightarrow \mathrm{L}, &
\text{if } \hat{s}^{K}(V_i,T_j)\le \hat{s}_{\mathrm{mean}} ~\text{and}~ |\delta_{i,j}|\le \epsilon, \\[0.6ex]
\mathrm{H}\rightarrow \mathrm{H}, &
\text{if } \hat{s}^{K}(V_i,T_j)> \hat{s}_{\mathrm{mean}} ~\text{and}~ |\delta_{i,j}|\le \epsilon, \\[0.6ex]
\mathrm{L}\rightarrow \mathrm{H}, &
\text{if } \hat{s}^{K}(V_i,T_j)> \hat{s}_{\mathrm{mean}} ~\text{and}~ \delta_{i,j}>\epsilon, \\[0.6ex]
\mathrm{H}\rightarrow \mathrm{L}, &
\text{if } \hat{s}^{K}(V_i,T_j)\le \hat{s}_{\mathrm{mean}} ~\text{and}~ \delta_{i,j}<-\epsilon.
\end{cases}
\end{equation}

\begin{figure}[t]
\includegraphics[width=\columnwidth]{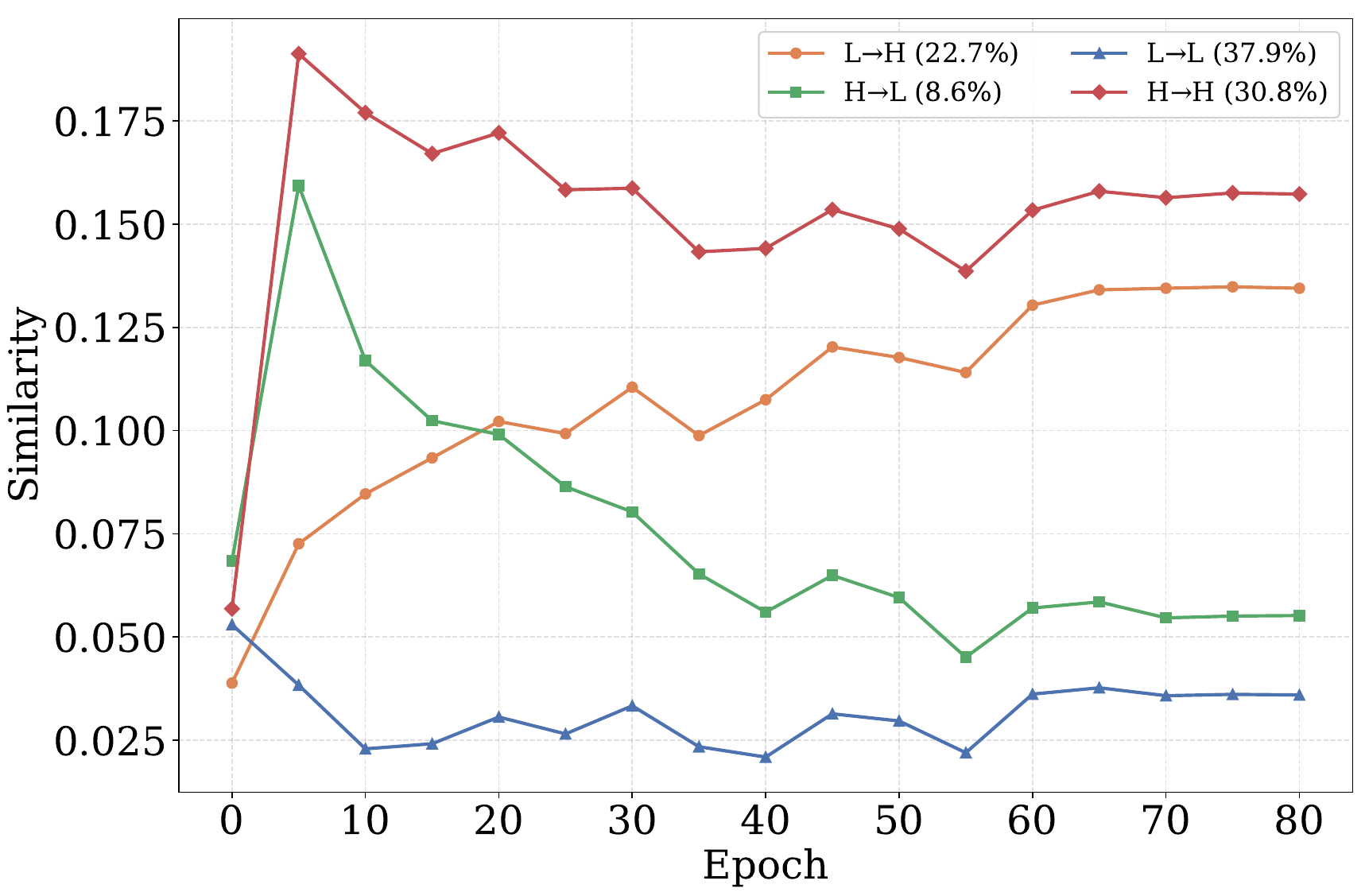}
\caption{The average cosine similarity curves for negative pair categories based on 
\label{fig:class_detail_re}
Section~\ref{sec_experiments} architecture}
% The legends denote the similarity trend of a pair during training: HH (remains high), HL (decreases), LL (remains low), and LH (increases).
\end{figure}

\section{Contrastive Learning Model Configuration.}\label{sec:CLmodel}
% To ensure architectural consistency between the Contrastive Learning Model (inference model) in Figure~\ref{fig:pipeline} and the subsequent SLT Model, we employ the Visual and Text Encoders described in Section~\ref{subsec:framework}. We conducted an analysis similar to the preliminary experiments in Section~\ref{sec_preliminary}, but with updated configurations: the model here comprises a 12-layer encoder optimized via LoRA, consistent with the settings in Section~\ref{sec_experiments}.We further analyzed the distribution of negative sample categories. As shown in Figure~\ref{fig:class_detail_re}, while the category proportions show minor deviations, the "well-behaved" negative samples (L~$\rightarrow$~L and H~$\rightarrow$~L) account for only 46.5\%. Notably, the distribution of similarity scores is mini, because the current model utilizes LoRA with a lower learning rate ($1 \times 10^{-4}$) to better preserve the LLM's linguistic capabilities. We utilize this configured model as the Contrastive Learning Model for calculating batch scores.
To ensure architectural consistency between the Contrastive Learning Model (inference model) in Figure~\ref{fig:pipeline} and the downstream SLT Model, we utilize the identical Visual and Text Encoders described in Section~\ref{subsec:framework}. We replicated the preliminary analysis from Section~\ref{sec_preliminary} using this updated configuration, specifically a 12-layer encoder optimized via LoRA, consistent with Section~\ref{sec_experiments}. As illustrated in Figure~\ref{fig:class_detail_re}, while the category proportions show minor deviations, the ``well-behaved'' negative samples (L~$\rightarrow$~L and H~$\rightarrow$~L) account for only 46.5\%. Notably, the distribution of similarity scores exhibits a narrower range. This is attributed to the use of LoRA with a reduced learning rate ($1 \times 10^{-4}$), which restricts the magnitude of feature updates to better preserve the LLM's linguistic capabilities. We employ this calibrated model to calculate batch scores for pair selection.

\begin{table}[t]
\centering
\small % 保持小字号，这很好
\setlength{\tabcolsep}{5pt}
% 1. 使用 tabularx，设置总宽为 \columnwidth
\begin{tabularx}{\columnwidth}{l Y Y Y Y Y} 
\toprule
% 注意：这里不需要 resizebox 了
\multirow{2}{*}{\textbf{Method}} & \multicolumn{5}{c}{\textbf{PHOENIX14T}}\\
\cmidrule(lr){2-6}
 & \textbf{R}  &  \textbf{B1} & \textbf{B2} & \textbf{B3} & \textbf{B4} \\
\midrule
% 2. 记得给 [cls] 加上花括号防止超时
BaseLine(End-to-End) & 37.38 & 38.37 & 28.52 & 22.27 & 18.16 \\
GFSLT-VLP$^\dagger$  & 27.89& 28.13 & 18.87 & 13.84 & 10.96  \\
GFSLT-VLP(Ours)   & 40.67 & 41.82 & 31.49 & 24.72 & 20.27\\

\bottomrule
\end{tabularx}
\caption{\textbf{Analysis of Contrastive Learning effectiveness.} We compare the baseline with GFSLT-VLP variants on PHOENIX14T. $\dagger$ denotes that the method was adapted to an end-to-end architecture for a fair comparison.}
\label{tab:cl_effectiveness}
\end{table}
\section{Impact of Contrastive Learning on SLT}
\label{sec:cl impaction}
To investigate the impact of Contrastive Learning (CL) on SLT, we reproduced the GFSLT-VLP~\cite{gloss-free-VLP} method on the PHOENIX14T dataset. For a fair comparison with our baseline, we adapted GFSLT-VLP into an end-to-end architecture (denoted by $\dagger$), where the visual encoder and translation decoder are trained jointly from scratch for 200 epochs.

As shown in Table~\ref{tab:cl_effectiveness}, directly incorporating a contrastive objective into the end-to-end SLT baseline results in a precipitous performance drop (e.g., BLEU-4 falls from 18.16 to 10.96). This observation aligns with the findings in Section~\ref{par:effect SCL}, where adding standard CL yielded negligible gains (+0.06 BLEU-4). Conversely, when employing CL as a pre-training stage (as in the original GFSLT-VLP and our SCL-SLT), it consistently boosts translation performance (reaching 20.27 BLEU-4). This suggests that CL is most effective when used to learn robust representations prior to the translation task, rather than as a simultaneous auxiliary loss in an end-to-end pipeline.

\end{document}